\documentclass[twocolumn,twoside,letterpaper,10pt]{IEEEtran}
\usepackage{cite}
\usepackage{pifont}
\usepackage{amsmath,amssymb,amsfonts}
\usepackage{graphicx}
\usepackage{textcomp}
\usepackage{xcolor}
\usepackage[ruled]{algorithm2e} 
\usepackage{algpseudocode}  
\usepackage{CJK}
\usepackage{listings}
\usepackage{bm,graphicx,multirow,multicol,bbm,subfigure,color,mdframed,wasysym,subeqnarray}
\usepackage{hyphenat}
\usepackage{amssymb}

\begin{document}

\title{Multi-Robot Object Transport Motion Planning with a Deformable Sheet}
\author{Jiawei~Hu, Wenhang~Liu, Heng~Zhang,	Jingang~Yi,~\IEEEmembership{Senior~Member,~IEEE,} and Zhenhua~Xiong,~\IEEEmembership{Member,~IEEE}
\thanks{Manuscript received: March 1, 2022, Accepted June 24, 2022.}
\thanks{This paper was recommended for publication by Editor Hyungpil Moon upon evaluation of the Associate Editor and Reviewers' comments.} 
\thanks{This work was supported in part by the National Natural Science Foundation of China (U1813224), Ministry of Education China Mobile Research Fund Project (MCM20180703) and MoE Key Lab of Artificial Intelligence, AI Institute, Shanghai Jiao Tong University, China.(\textit{Corresponding author: Zhenhua~Xiong, Jingang~Yi}).}

\thanks{J. Hu, W. Liu, H. Zhang, and Z. Xiong are with the School of Mechanical Engineering, Shanghai Jiao Tong University, Shanghai, China.  {\tt\footnotesize  hu\_jiawei@sjtu.edu.cn; liuwenhang@sjtu.edu.cn; zhanghengme\_sjtu@sjtu.edu.cn; mexiong@sjtu.edu.cn}} \thanks{J. Yi is with the Department of Mechanical and Aerospace Engineering, Rutgers University, Piscataway, NJ 08854 USA. {\tt\footnotesize  jgyi@rutgers.edu}.}
\thanks{Digital Object Identifier (DOI): see top of this page.}
}

\markboth{IEEE Robotics and Automation Letters. Preprint Version. Accepted June, 2022}
{Hu \MakeLowercase{\textit{et al.}}: Multi-Robot Object Transport Motion Planning with a Deformable Sheet} 

\maketitle

\begin{abstract}

Using a deformable sheet to handle objects is convenient and found in many practical applications. For object manipulation through a deformable sheet that is held by multiple mobile robots, it is a challenging task to model the object-sheet interactions. We present a computational model and algorithm to capture the object position on the deformable sheet with changing robotic team formations. A virtual variable cables model (VVCM) is proposed to simplify the modeling of the robot-sheet-object system. With the VVCM, we further present a motion planner for the robotic team to transport the object in a three-dimensional (3D) cluttered environment. Simulation and experimental results with different robot team sizes show the effectiveness and versatility of the proposed VVCM. We also compare and demonstrate the planning results to avoid the obstacle in 3D space with the other benchmark planner.
\end{abstract}



\section{Introduction}
\label{sec1}

\IEEEPARstart{D}{eformable} sheets are commonly used as flexible carriers in many object manipulation applications, such as transferring patients with bed sheets in hospital~\cite{RoyTASE2005}. Due to sheet deformability, multiple supporters (e.g., mobile robots) are needed to carry the object in the sheet. For robot-sheet-object transport systems, sheet deformability provides an advantage of manipulating objects by changing the relative positions among the robotic team members. However, this feature also brings challenges for real-time modeling of the object-sheet interactions, robot formation control and motion planning. The goal of this paper is to present an efficient computational approach for multi-robot motion planning for manipulating and transporting an object in a deformable sheet held by the robotic team.   

Using a mobile robot team to collaboratively transport or manipulate a rigid object has been reported in previous decades. By different contact methods, multi-robot manipulation (MRM) can be classified into three types: pushing, grasping, and caging~\cite{tuci2018cooperative}. In these systems, robots are in direct contact with the transported or manipulated objects. The work in~\cite{wang2016force} proposed a distributed multi-mobile robot control method to transport a rigid object. The fixed formation shape did not allow the robot team to adjust relative positions between any two robots for obstacle avoidance in cluttered environment, such as narrow passes. For multi-mobile manipulators, Ren {\em et al.}~\cite{ren2020fully} presented a fully distributed control scheme for a team of networked mobile manipulators to transport an unknown object. The manipulator is expensive and its kinematic redundancy also brings additional planning and control design challenges. 

Using deformable sheets to hold an object provides a new way for MRM. The motion of objects on the sheet is highly dependent on the robot motion and is not easily predicted due to high dimensionality of deformable sheets~\cite{herguedas2019survey}. Physical and geometric models are among the main modeling methods of the deformable sheet~\cite{nadon2018multi}. Physical models such as the mass-spring model~\cite{provot1995deformation} and finite element method~\cite{weil1986synthesis} are common in computer graphics. The work in~\cite{bai2016dexterous} studied a set of cloth manipulation tasks in daily activities, including folding laundry, wringing a towel, and putting on a scarf. The motion between the rigid body and the cloth was calculated by a physics-based simulator, which involves complex meshing and calculations and cannot be used for real-time applications.

In robotic manipulation, geometric models are used to reduce the high dimensionality of the sheet. In~\cite{alonso2015local, chand2019transportation}, the geometric constraints of the deformable sheet are used for motion planning without the complete model. The work in~\cite{mcconachie2020manipulating} used a set of discrete points to approximate the sheet state. Assuming a completely flexible sheet, the work in~\cite{hunte2019collaborative,HunteMECC2021} established the geometric link model of the sheet-object kinematic relationship with a three-robot team. The recent work in~\cite{HunteMECC2022} extended the sheet-object kinematic model to include rotational motion of a spherical-shape object for pose manipulation. As the number of robots increases, the transported object may have different equilibrium states on the sheet, and the the physical principle-based kinematic models developed in~~\cite{HunteMECC2021,HunteMECC2022} becomes complicated and difficult to obtain precisely. 

Inspired by the configuration of cable suspended robots~\cite{capua2009motion, capua2010motion, capua2011spiderbot}, we propose the virtual variable cables model (VVCM) that simplifies the object-sheet interactions as a set of  dynamically changing cables to collaboratively hold the object. The equilibrium of the object on the sheet is then represented by the tightness status of virtual cables in the quasi-static state. A computational algorithm is proposed for obtaining real-time object position in the sheet that is held by arbitrary numbers of mobile robots. Based on the VVCM, we also propose a motion planner for the robotic team for manipulating and transporting the object in cluttered environment. It is difficult to use the artificial potential field method~\cite{hu2020distributed} or sampling-based planning algorithms~\cite{zhang2021task} to maintain the shape of the multi-robot formation. Alonso {\em et al.}~\cite{alonso2017multi} computed the optimal robotic formation through a constrained nonlinear optimization problem for multi-mobile manipulator-base object manipulation with the deformable sheet. The motion planners in~\cite{alonso2015local,chand2019transportation,mcconachie2020manipulating,hunte2019collaborative,HunteMECC2021,alonso2017multi}  did not consider the three-dimensional (3D) position of the transported object with deformable sheets. For obstacle avoidance, the object is always planned to bypass the obstacle, that is, planar motion in two-dimensional (2D) space. Our planner is an extension of these planners and we exploit the possible obstacle avoidance in 3D space by allowing the object to move above the obstacles, which is referred as obstacle crossing. Therefore, the proposed motion planner allows the robotic team to avoid obstacles in both 2D or 3D space, which is efficient and effective in cluttered environments (e.g., a narrow corridor with obstacles). 

The main contributions of this paper are twofold. First, the proposed VVCM for robots-sheet-object interactions is new and provides an efficient method to compute the different equilibrium of the object on the sheet held by arbitrary number of robots. Second, the motion planning algorithm extends the existing planners for MRM with the new capability of obstacle crossing, which not only expands the workspace of the robot team, but also improves the efficiency.

The remainder of this paper is organized as follows. We first present the problem statement in Section~\ref{sec2}. Section~\ref{sec3} presents the computational model for object manipulation with a deformable sheet. Section~\ref{sec4} discusses the motion planner with obstacle crossing capability. Section~\ref{sec5} presents the experimental results and finally, Section~\ref{sec6} summarizes the concluding remarks.

\section{System Configuration and Problem Statement}
\label{sec2}

\subsection{System Configuration}
\label{sysconfig}

\begin{figure}[h!]
	\vspace{-1mm}
\centering
	\includegraphics[width=0.98\columnwidth]{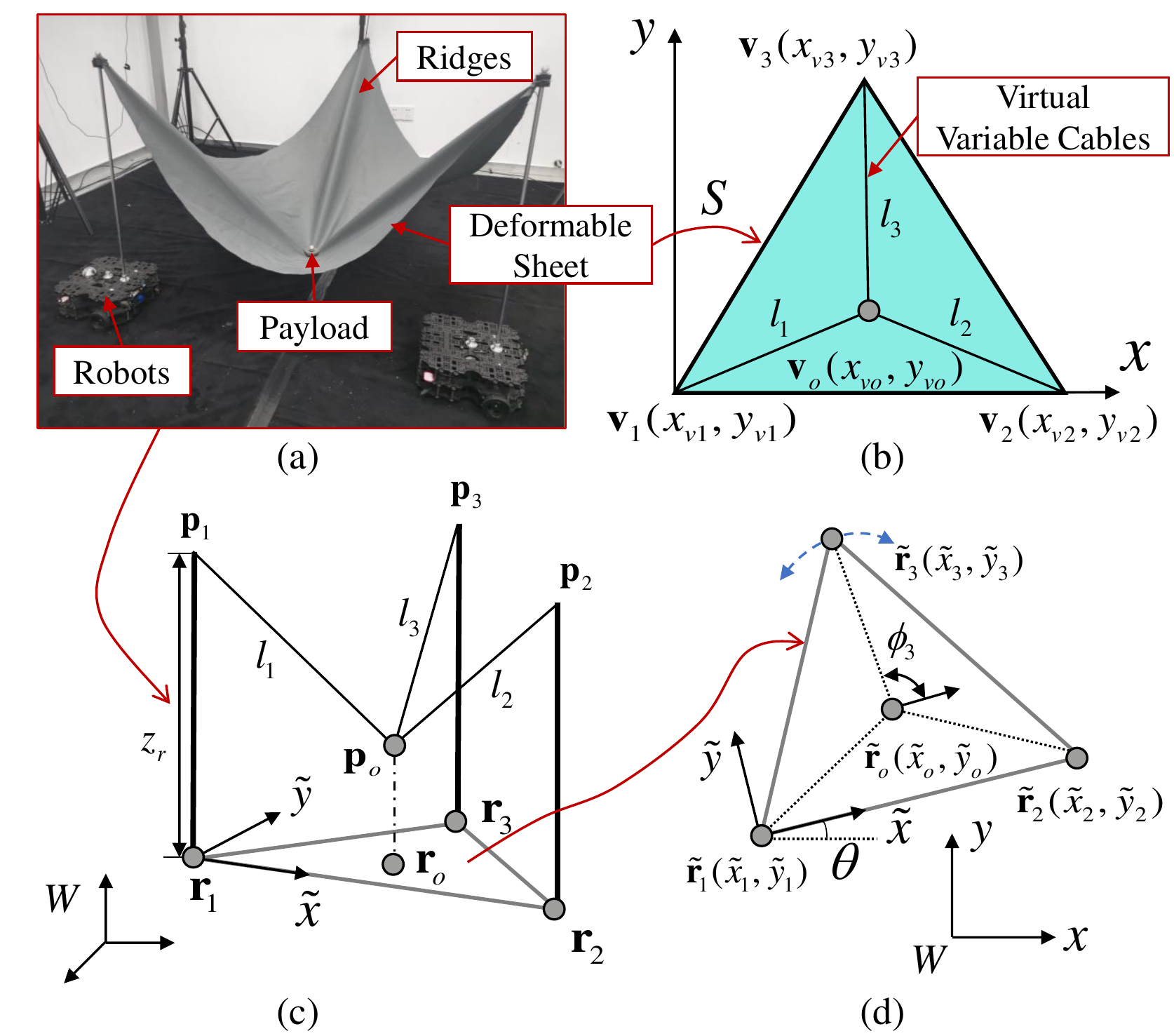}
	\centering
	\caption{A three-robot team handles a payload held by a deformable sheet. (a) System configuration introduction. (b) Local coordinate frame $\mathcal{S}$ along the non-deformed sheet. (c) Illustration of the virtual variable cables model. (d) The projected robot formation on the base and its local coordinate system (top view). Note that, $\phi_i$ can vary within the geometric constraints of the sheet and results in different formations.}	
	\label{fig_config}       
\end{figure}

We consider the object manipulation and transporting by a deformable sheet that is held by $N$ mobile robots, $N\in \mathbb{N}$. Fig.~\ref{fig_config}(a) illustrates the system configuration for a three-robot case, that is, $N=3$. The trajectory of the transported object, denoted as $O$, is in 3D space and the mobile robots only move on the planar surface.  A deformable sheet is held by an $N$-mobile robot team and the robotic holding point $\bm{p}_i$, $i=1,\ldots,N$, is fixed at a constant height. 

To describe the motion of object $O$ on the deformable sheet, we consider the local sheet frame, denoted as $\mathcal{S}$, on the plat sheet plane. The coordinate of points $\bm{p}_i$ on $\mathcal{S}$ is denoted as $\bm{v}_{i}=[x_{vi}\;y_{vi}]^T$, $i=1,\ldots,N$. The contact point between object $O$ and the sheet is denoted as $\bm{v}_{o}=[x_{vo}\;y_{vo}]^T$ in $\mathcal{S}$, as shown in Fig.~\ref{fig_config}(b). A world coordinate frame $\mathcal{W}$ is also used and the coordinate of $\bm{p}_i$ is denoted as $\bm{p}_i=[\bm{r}_i^T \; z_r]^T$, $\bm{r}_i=[x_i \; y_i]^T$, $i=1,\ldots ,N$, where $z_r$ is a constant height for all robots. The coordinate of object $O$ in $\mathcal{W}$ as $\bm{p}_o= [x_o\;y_o\;z_o]^T$. Fig.~\ref{fig_config}(c) further illustrates the coordinates of $\bm{p}_i$ and $\bm{p}_o$. The collection of all robots' position is denoted as $\mathcal{R}_N=\{\bm{r}_1,\cdots,\bm{r}_N\}$.  

The motion of the object $O$ is closely related to the robot motion and the material and shape of the sheet. To precisely describe the motion and restrict the problem scope, we use the following considerations and assumptions. First, the object $O$ is assumed to be a mass point. Second, the object transport and motion are quasi-static, that is, the robot team moves slowly and smoothly and the dynamic effects of object $O$'s motion are negligible. Finally, the deformable sheet is completely soft and inelastic and thus, the object $O$ moves freely on the sheet under gravity. With these considerations, the sheet holding points $\bm{v}_i$ in $\mathcal{S}$ form a convex polygon with $N$ vertices and the object $O$ is located inside the polygon; see Fig.~\ref{fig_config}(b) for a triangle case (i.e., $N=3$). It is clear that in non-constrained conditions, the object $O$ tends to move and stay at the position of minimum potential energy.

\begin{figure}[h!]
	\includegraphics[width=\columnwidth]{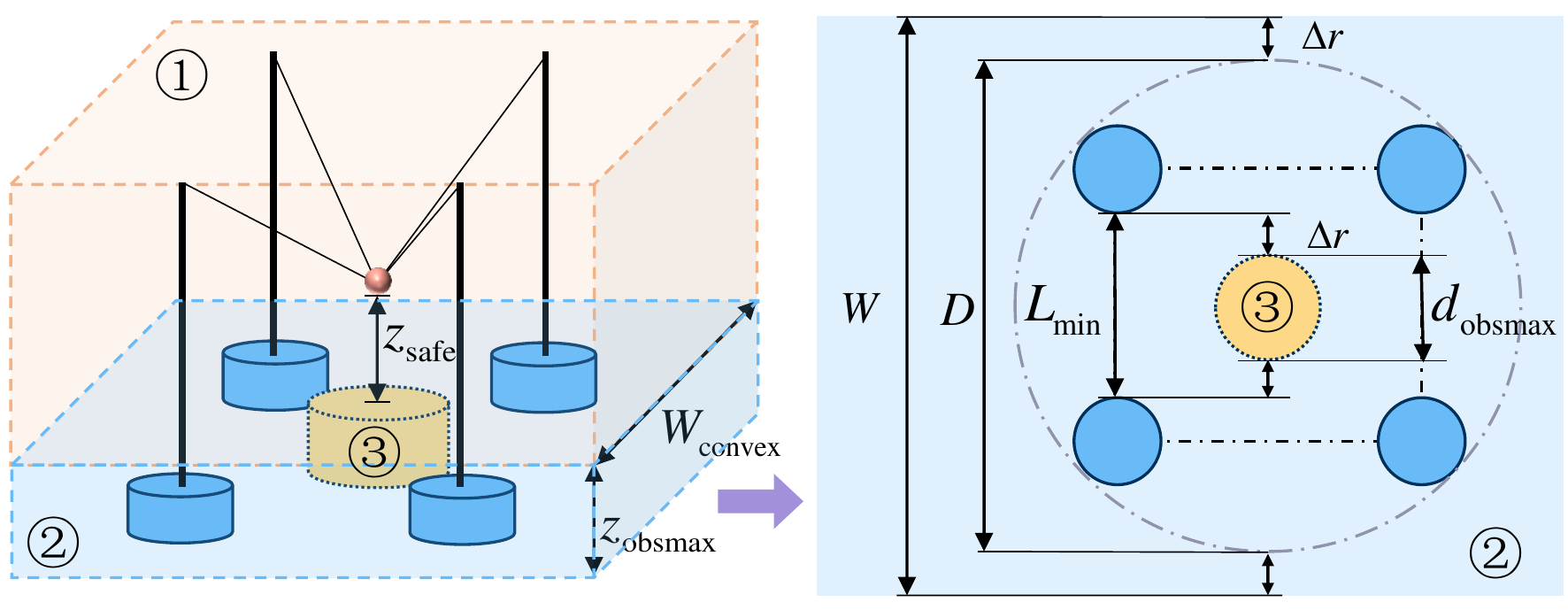}
	\centering
	\caption{Illustration of the multi-robot formation in 3D space and description of the formation properties. Left: side-view of the 4-robot team with circular base and a circular obstacle. Right: Top-view of the 4-robot team with the obstacle. Volume \ding{172} indicates obstacle-free convex region, \ding{173} indicates the multi-robot formation region, and \ding{174} represents the traversable obstacle region.}
	\label{fig_indicators}       
\vspace{-2mm}
\end{figure}

The 3D position of object $O$ is directly related to the shape of the robot formation. Fig.~\ref{fig_indicators} illustrates the multi-robot formation configuration. The object is considered to be transported by the robotic team in a space with width $W_\text{convex}$; see the left plot of Fig.~\ref{fig_indicators}. To quantify the formation properties, we introduce a few geometric variables that will be used later in motion planning design, including the robotic formation size $W$, the maximum traversable obstacle height $z_{\text{obsmax}}$ and the maximum traversable obstacle diameter $d_{\text{obsmax}}$.

The robotic formation size $W$ is defined as 
\begin{equation}
	\label{eq_W}
	W = D + 2\Delta r
\end{equation}
where $D$ is the diameter of the minimum circumscribed circle of the robotic team and $\Delta r$ is a constant safety distance at each side of the robotic formation; see Fig~\ref{fig_indicators}. The maximum traversable obstacle height is obtained as
\begin{equation}
	\label{eq_zobs}
	z_\text{obsmax}=z_o-z_{\text{safe}}
\end{equation}
where $z_{\text{safe}}$ is a minimum safety distance between the object and the obstacle. We assume that each robot is a particle and we define the shortest distance between any two robots as 
\begin{equation}
L_{\min}= \min_{i,j\in\bm{I}_N, i\neq j} \| \Delta  \bm{r}_{ij} \|, \, \Delta  \bm{r}_{ij}=\bm{r}_{i} - \bm{r}_j
\end{equation}
where index set $\bm{I}_N=\{1, \ldots, N\}$. The maximum traversable obstacle diameter $d_{\text{obs}}$ is then defined as 
\begin{equation}		
	d_{\text{obsmax}}=L_{\min}-2\Delta r.
	\label{eq_dobs}
\end{equation}
From~\eqref{eq_zobs}-\eqref{eq_dobs}, variables $W$, $d_{\text{obsmax}}$ and $z_{\text{obsmax}}$ are the functions of formation shape that is determined by $\mathcal{R}_N$ and we will use these variables to design motion planner in Section~\ref{sec4}.

\subsection{Problem Statement}

One important observation of the object $O$ in the deformable sheet that is held by $N$ points is that the connection between $\bm{p}_i$ and $\bm{p}_o$ in frame $\mathcal{S}$ can be a straight-line (i.e., full tension) or a curved-line (i.e., slack). We call the connected line between $\bm{p}_i$ and $\bm{p}_o$ on $\mathcal{S}$ as {\em ridge}. By the local coordinates in frame $\mathcal{S}$, the geodesic length of the $i$th ridge is denoted as $l_i=\|\bm{v}_i-\bm{v}_o\|$, $i=1, \ldots,N$. The planar position of the $i$th robot is captured by $\bm{r}_i$. We assume that the initial position of all robots and obstacles in 3D space are known. The problem statement is given as follows. 

{\em Problem Statement}: The motion planning problem is to design the trajectory of robot team $\mathcal{R}_N$ such that the object $O$ can be safely transport from the initial to target positions in 3D space. 

\section{Multi-Robot Manipulation Models}
\label{sec3}

In this section, we present a computational approach to obtain the object position under the motion of the robotic team. The goal of the multi-robot manipulation model is to compute object positions $\{\bm{p}_o,\bm{v}_o\}$ for given $\mathcal{R}_N$. 

\subsection{Virtual Variable Cables Model (VVCM)}
\label{sec3_A}

Due to the inelastic property, the sheet deformation is distance-preserving. As the robot formation changes, the object $O$ might move on the sheet under gravity and therefore, the contact point $\bm{v}_o$ also changes. Since the geodesic distance $l_i$ between $\bm{v}_o$ and $\bm{v}_i$ on $\mathcal{S}$ varies, the $i$th ridge can be viewed as  a virtual cable connecting points $\bm{p}_i$ and $\bm{p}_o$; see Fig.~\ref{fig_config}(c). Therefore, the object $O$ can be regarded as being transported by $N$ variable-length cables that are fixed on the robot. Because the virtual cable might be taut or slack, it is clear that the Euclidean distance between object $O$ and vertex $\bm{p}_i$ is less than or equal to the corresponding virtual cable length, namely,
\begin{equation}
	\label{ineq_rope}		
	 l_i = \| \bm{v}_o -\bm{v}_i \| \geq \| \bm{p}_o - \bm{p}_i \|,  \; i=1, \dots, N.
\end{equation}

The complexity of the above described virtual cable-object interactions lies in the unspecified taut/slack status of each cable under robot formation. The positions $(\bm{v}_o, \bm{p}_o)$ of the object $O$ are jointly determined by the set of taut cables. To precisely determine the object position, we define the status of the $i$th virtual cable by variable $I_i=0$ if the cable is slack; otherwise, $I_i=1$ when it is taut, $i \in \bm{I}_N$. Furthermore, defining taut cable index set $\mathbb{I}_t=\{i: I_i=1, i\in\bm{I}_N\}$, we denote $m=|\mathbb{I}_t| \leq N$ as the cardinality of $\mathbb{I}_t$. When the $i$th cable is taut (i.e., $I_i=1$), it satisfies
\begin{equation}
	\label{eq_rope}
		l_i = \| \bm{v}_o -\bm{v}_i \| = \| \bm{p}_o - \bm{p}_i \|
\end{equation}
The direct kinematics is to know the robotic formation $\mathcal{R}_N$ to solve the positions $(\bm{v}_o, \bm{p}_o)$ of the object $O$. To obtain an algebraic solution, the formation can be transformed into a local coordinate system, as shown in Fig.~\ref{fig_config}(d). In the local coordinate system, with ${\bm{r}}_1$ as the origin and $\overrightarrow{{\bm{r}}_1{\bm{r}}_2}$ as the $x$-axis, the robots' coordinates are arranged in counterclockwise order, denoted as $\tilde{\bm{r}}_i = [\tilde{x}_i\;\tilde{y}_i]^T, \, i\in\bm{I}_N$. The projected coordinate of the transported object $O$ is denoted as $\tilde{\bm{r}}_o = [\tilde{x}_o\;\tilde{y}_o]^T$ in the local frame. The direct kinematics can be simplified as that given $\tilde{\bm{r}}_i$, $i\in\bm{I}_N$, and $\tilde{x}_1 = \tilde{y}_1 =\tilde{y}_2=0$ by the above setup, we need to find $\bm{v}_o$ and $(\tilde{\bm{r}}_o, z_o)$, which consists of five independent coordinate variables of the transported object.

\begin{figure}[h!]
	\centering
	\includegraphics[width=\columnwidth]{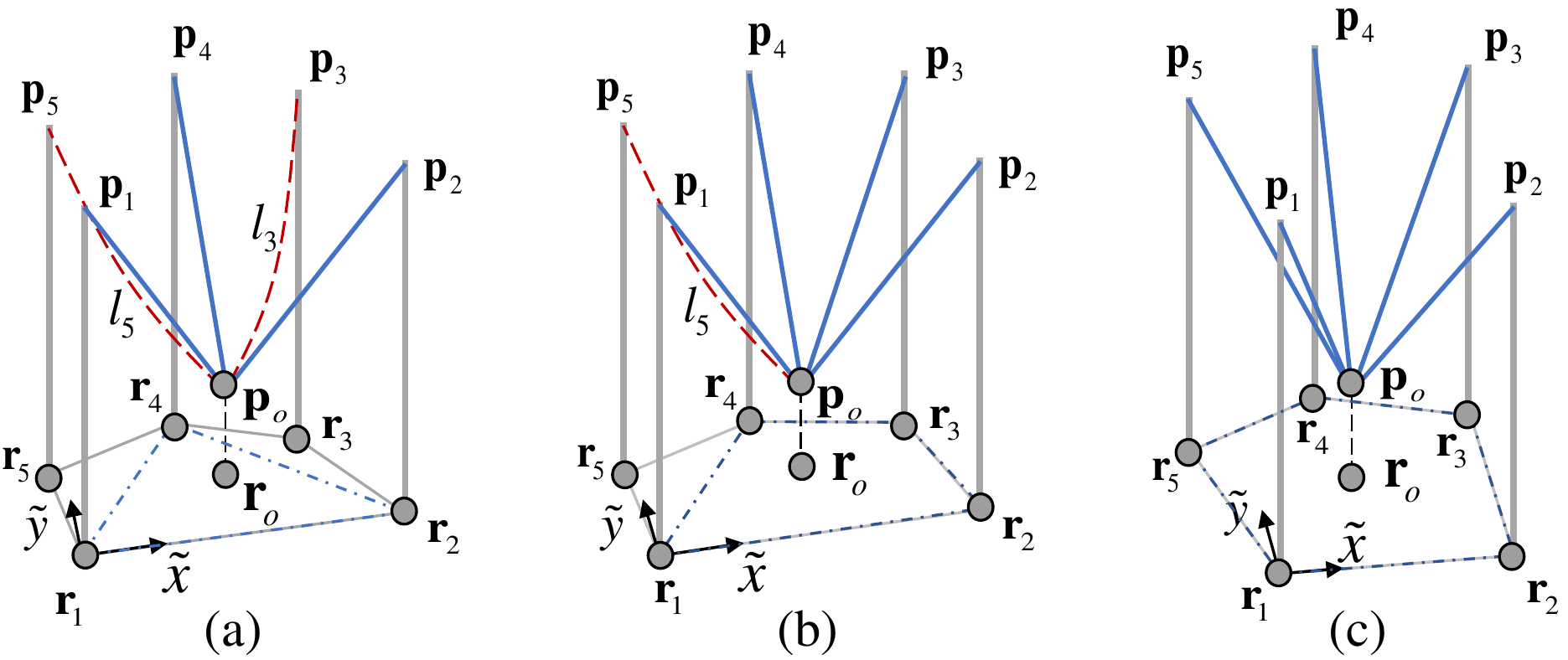}
	\caption{Three static equilibrium conditions of Virtual Variable Cables Model. (a) $\bm{p}_o$ is determined by the triangle sub-formation. (b) $\bm{p}_o$ is determined by the quadrilateral sub-formation. (c) $\bm{p}_o$ is determined by the pentagon formation.}
	\label{fig_vvcm}       
\end{figure}

For the VVCM with an $N$-robot team, different combinations of the taut virtual cables represent different geometric constraints. Because of the above-mentioned five independent variables, we use 5-robot team as an illustrative example to determine possible equilibrium conditions of the transported object. Fig.~\ref{fig_vvcm} illustrates three possible equilibrium state of the object $O$ under a five-robot team. Among five virtual cables, the numbers of taut cables can be 3, 4, or 5 as illustrated in Fig.~\ref{fig_vvcm}(a)-(c), respectively.  

The projection of $\bm{p}_i$, $i \in \mathbb{I}_t$, on the local frame must be convex polygons; otherwise, the $i$th virtual cable must be slack, that is, $i \notin \mathbb{I}_t$, which results in a change of the object's equilibrium state. We first consider the case when five cables are taut, i.e., $m=5$, as shown in Fig.~\ref{fig_vvcm}(c). In this case, all five cables satisfy the geometric constraints~\eqref{eq_rope}. Through these five independent equations, the required five independent variables can be obtained. The calculation function is denoted as \emph{VVCM-Pentagon} as shown in the computational approach with pseudo-code illustrated in Algorithm~\ref{Al_DK}. When more than five cables are taut in a larger robot team, $m>5$, the object is constrained by full redundancy and its position is determined by any five independent equations in the geometric constraints~\eqref{eq_rope}. Therefore, we can obtain $(\bm{p}_o,\bm{v}_o)$ by selecting five taut cables from the team.

Fig.~\ref{fig_vvcm}(a) shows the case when two cables are slack and the object's position is determined by the triangle sub-formation. The configuration is considered as a three-robot team, $m=3$, similar to that as shown in Fig.~\ref{fig_config}. In this case, combining the three geometric constraint equations~\eqref{eq_rope} in the local coordinate system; see Fig.~\ref{fig_config}(d), we obtain 
\begin{eqnarray}
    \tilde{x}_o&=&\frac{x_\text{v2}}{\tilde{x}_2}x_{vo}+\frac{\tilde{x}_2^2-x_\text{v2}^{2}}{2\tilde{x}_2}, \nonumber\\   
\tilde{y}_o&=&\left(\frac{x_\text{v3}}{\tilde{y}_3}-\frac{\tilde{x}_3}{\tilde{y}_3}\frac{x_\text{v2}}{\tilde{x}_2} \right)x_{vo}+ \frac{y_\text{v3}}{\tilde{y}_3}y_{vo}-\frac{\tilde{x}_3}{\tilde{y}_3}\frac{\tilde{x}_2^2-x_\text{v2}^2}{2\tilde{x}_2} \nonumber \\
&&+\frac{\tilde{x}_{3}^2+\tilde{y}_3^2-x_\text{v3}^2-y_\text{v3}^2}{2\tilde{y}_3}.	
	\label{eq_xy}
\end{eqnarray}
Two additional equations are needed to determine $\bm{v}_o$. A convex optimization problem is formulated to minimize $J_z(\bm{v}_o)= -(z_o - z_r)^2 = \tilde{x}_o^2 + \tilde{y}_o^2 - x_{vo}^2 - y_{vo}^2$ for obtaining $\bm{v}_o$ by the observation that the object should rest at the lowest potential energy, namely,  
\begin{equation}
	\label{opt_Jz}
	\bm{v}_o^*= \arg \min_{\bm{v}_o} \ J_z (\bm{v}_o), \ {\bm{v}_o \in \bigtriangleup\bm{v}_{1} \bm{v}_{2}\bm{v}_{3} \subset \mathcal{S}},
\end{equation}
where $\bigtriangleup\bm{v}_{1} \bm{v}_{2}\bm{v}_{3}$ represents the triangle formed by vertices $\bm{v}_{1}$, $\bm{v}_{2}$, and $\bm{v}_{3}$; see Fig.~\ref{fig_config}(b). The convex polygon formation assumption of the robot team allows us to solve~(\ref{opt_Jz}) by convex optimization method. From the first-order necessary condition of~(\ref{opt_Jz}), we obtain the gradient $\nabla J_z(\bm{v}_o)=\bm{0}$ and it reduces to 
\begin{equation}		
	\label{eq_vo}	
 \left\{ \begin{matrix}
		a_{11}{x}_{vo} + a_{12}{y}_{vo} = b_1 \\
a_{21}{x}_{vo} + a_{22}{y}_{vo} = b_2
	\end{matrix} \right. 
\end{equation}
where $a_{11} = \frac{x_\text{v2}^2-\tilde{x}_2^2}{x_\text{v2}}$, $a_{12} = \frac{\tilde{x}_2 (x_\text{v3}\tilde{x}_2-{\tilde{x}_3}{x_\text{v2}})}{x_\text{v2}y_\text{v3}}$, $b_1 = \frac{x_\text{v2}^2-\tilde{x}_2^2}{2}$, $a_{21}=x_\text{v3} \tilde{x}_2 - \tilde{x}_3 x_\text{v2}$, $a_{22} = \frac{\tilde{x}_2 (y_\text{v3}^2 - \tilde{y}_3^2)}{y_\text{v3}}$, $b_2 = \frac{\tilde{x}_3 (\tilde{x}_2^2-x_\text{v2}^2)}{2} + \frac{\tilde{x}_2 (x_\text{v3}^2 + y_\text{v3}^2 - \tilde{x}_3^2-\tilde{y}_3^2)}{2}$. Thus, $\bm{v}_o=[x_{vo} \; y_{vo}]^T$ is obtained by solving~\eqref{eq_vo} and then $\tilde{\bm{r}}_o$, $z_o$ are obtained by~\eqref{eq_xy} and~\eqref{eq_rope} respectively. It is straightforward to verify that the Hessian matrix of $J_z$ is positive definite and therefore, the obtained optimal resolution is the global minimum. $\bm{r}_o$ is then obtained by coordinate transformation from $\tilde{\bm{r}}_o$, and finally, ${\bm{p}}_o$ is obtained. The calculation process is denoted as \emph{VVCM-Triangle} in Algorithm~\ref{Al_DK}.

For the last case with four taut virtual cable as shown in Fig.~\ref{fig_vvcm}(b), $m=4$, and the calculation process is similar to that of $m=3$ as described above. We omit the details and the process is denoted as \emph{VVCM-Quadrilateral} in Algorithm~\ref{Al_DK}. Through the VVCM, the position of the object $O$ is obtained and the results depend on the taut or slack status of the virtual cables. In the simple case of three-robot configuration, the direct kinematics solution is obtained from~\eqref{eq_vo}. Algorithm~\ref{Al_DK} summarizes the above results to compute the object position ($\bm{p}_o, \bm{v}_o$) manipulated by the $N$-robot team.

\begin{algorithm}[ht!]
	\vspace{1mm}
	\SetAlgoVlined
	\label{Al_DK}
	\caption {Deformable sheet object computation}
	\SetKwInOut{Input}{Input}
	\SetKwInOut{Output}{Output}
	\SetKwFunction{Linecross}{Linecross}
	\SetKwFunction{Polygon}{Polygon}
	\Input{$\mathcal{R}_N$, $z_r$, and $\{I_i\}_{1}^N$}
	\Output{$(\bm{p}_o,\bm{v}_o)$}
	 \eIf{$\emph{FormationFeasible}$}{
		 $m = 0, \, \mathbb{I}_t = \emptyset$\;
		 \For{$i=1$ to  $N$}{
		\lIf{$I(i) = 1$}{$m \leftarrow m + 1, \,  \mathbb{I}_t \gets  \{\mathbb{I}_t, i\}$}	}
	 \lIf{$m \geq 5$}{$(\bm{p}_o,\bm{v}_o) \gets \emph{VVCM-Pentagon} (\mathbb{I}_t,\mathcal{R}_N)$} 
	 \uElseIf{$m=4$}{$(\bm{p}_o, \bm{v}_o) \gets \emph{VVCM-Quadrilateral} (\mathbb{I}_t,\mathcal{R}_N)$}
 \ElseIf{$m=3$}{$(\bm{p}_o, \bm{v}_o) \gets \emph{VVCM-Triangle} (\mathbb{I}_t,\mathcal{R}_N)$}	
 \Return  $(\bm{p}_o, \bm{v}_o)$ 
}{
 \Return  $\textit{FALSE}$
} 
\end{algorithm}

\subsection{Inverse Kinematics}
\label{sec3_B}

To plan the robot motion, we need to consider the inverse kinematics and obtain $\mathcal{R}_N$ for the $N$-robot team with given $(\bm{p}_o,\bm{v}_o)$. When $\bm{v}_o$ is known, with the initial vertex of the deformable sheet, the length of each virtual cable $l_i$ is obtained. From the discussion in the previous section, the robotic team can have multiple solutions of $\mathcal{R}_N$ given different combinations of taut/slack status of the virtual cables.  

Algorithm~\ref{Al_DK} computes the position of the transported object under different numbers of robots and formations. However, not all formations are effective and efficient for transporting the object. As mentioned in the VVCM, once a virtual cable is slack, the corresponding robot no longer actively participates in handling the object, which leads to low efficiency. In addition, slack virtual cables can also cause system instability~\cite{capua2010motion}. Therefore, during transport, we assume that every virtual cable should be in taut all time and therefore satisfies~\eqref{eq_rope}. When the number of robots is less than five, i.e., $N<5$, $m<5$, the geometric constraints are not enough and the gravity constraint ~\eqref{opt_Jz} needs to be combined. Based on such observation of all-taut cables, the robot position are simply computed as
\begin{equation}
\bm{r}_i = \bm{r}_o + \sqrt{l_i^2 - (z_r - z_o)^2} \begin{bmatrix} \cos \phi_i \\ \sin \phi_i \end{bmatrix}
\label{eq_ik}
\end{equation}
with $\| \bm{r}_i - \bm{r}_j \| < \| \bm{v}_i -\bm{v}_j \|$, $i, j \in \bm{I}_N$ and $i \neq j$, where $\phi_i$ represents the rotation angle of the projection of the $i$th cable in the horizontal plane; see Fig.~\ref{fig_config}(d). Note that, $\phi_i$ can vary within the geometric constraints of the sheet and results in different formations. Based on the VVCM, three robots form the smallest size of robotic team to sufficiently and stably manipulate the spatial position of the object. We will discuss how to obtain the optimal robotic formation in~\eqref{eq_ik} through a multi-objective optimization method in the next section.

\section{Motion Planner with Obstacle Crossing Capability}
\label{sec4}

In this section, we apply VVCM to motion planning for multi-robot team to transport the object in cluttered environment. The planner first determines the robotic formation that is specified by relative positions among robots and then specifies the trajectory of the robotic team for crossing obstacles in 3D space.   

\subsection{Optimal Formation Generation} 
\label{sec4_A}

\begin{figure}[h!]
	\includegraphics[width=\columnwidth]{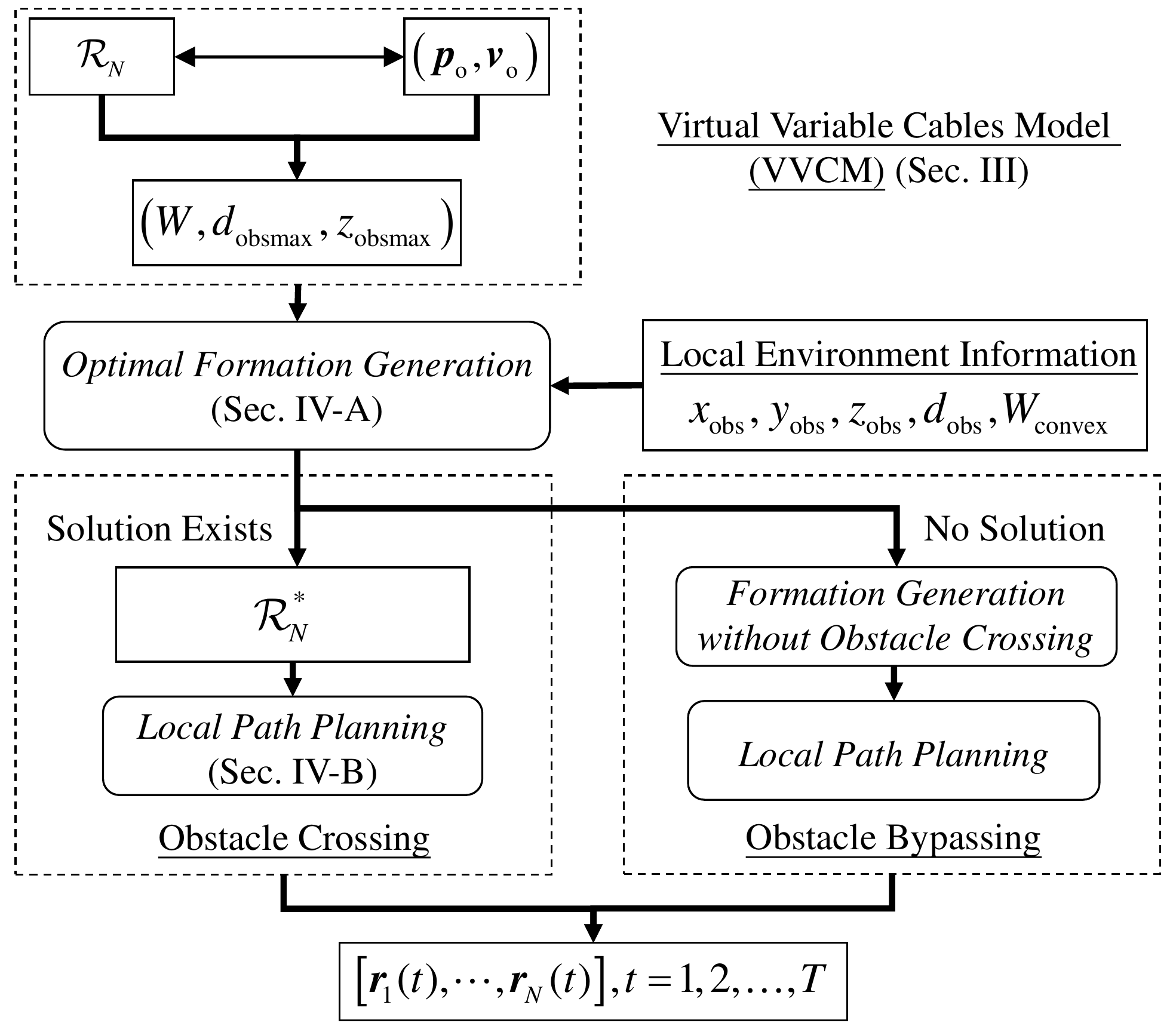}
	\centering
	\caption{The motion planner with obstacle crossing capability. }
	\label{fig_frame}       
\end{figure}

As discussed in Section~\ref{sec3_B}, the robot team has multiple possible formations due to combination of various taut/slack virtual cables and variable angle $\phi_i$. We formulate a constrained nonlinear optimization to compute an optimal robot formation. For obstacle avoidance in 3D space, passing from the top of obstacles (i.e., obstacle crossing) is preferred than passing from aside of obstacles. If no feasible formation is found for obstacle crossing, the planner then finds paths that bypass obstacles as demonstrated in~\cite{alonso2017multi}. Fig.~\ref{fig_frame} illustrates the overall design of the motion planner. 

The robot formation can be characterized by $\mathcal{R}_{\Delta}=\{\Delta  \bm{r}_{ij}\}$, $i,j\in\bm{I}_N, i\neq j$. The optimization problem considers to use formation variables to minimize cost ${J}_{f}$ that is given by
\begin{equation}		
J_f(\mathcal{R}_{\Delta})=J_{\text{trans}}+J_{\text{pass}}+J_{\text{cross}},
\label{eq_Jf}
\end{equation}
where $J_{\text{trans}}$, $J_{\text{pass}}$, and $J_{\text{cross}}$ are costs to penalize transportation, passable area and obstacle crossing, respectively. $J_{\text{trans}}$ ensures the safety and stability of system during transportation. In order to ensure that each robot participates in the handling of the object, each virtual cable is taut and $J_{\text{trans}}$ is given by
\begin{equation}	
J_{\text{trans}}=\lambda_1\|\bm{v}_o - \bm{v}_o^0\|^2 +\lambda_2\sum_{i \neq j}\left(\|\Delta  \bm{r}_{ij}\|-\| \Delta  \bm{r}^0_{ij}\|\right)^2
\label{eq_Jtrans}
\end{equation}	
where $\bm{v}_{o}^{0}$ and $\bm{r}_{i}^{0}$ are initial positions of the object $O$ in $\mathcal{S}$ and the $i$th robot's planar position, respectively, $\Delta  \bm{r}^0_{ij}=\bm{r}_i^0-\bm{r}_j^0$, and $\lambda_1,\lambda_2>0$ are the weighting factors. Cost $ J_{\text{pass}}$ keeps the robot team in a safe range and ensures that the formation can rotate freely during obstacle avoidance. Therefore, $ J_{\text{pass}}$ is designed as
\begin{equation}	
	J_{\text{pass}} = -{\lambda_3}(W_\text{convex} - W)^2,
 \label{eq_Jpass}	
\end{equation}	
where $\lambda_3>0$ penalizes the outline size of the robot team.  Cost $J_{\text{cross}}$ ensures that the formation can cross the obstacle in any direction and keeps the object at a safe height, 
\begin{equation}	
J_{\text{cross}} =-\lambda_4(z_{\text{obsmax}}-z_{\text{obs}})^2-\lambda_5(d_{\text{obsmax}}-d_{\text{obs}})^2
\label{eq_Jcross}
\end{equation}
where weights $\lambda_4,\lambda_5>0$ penalize the height of object crossing the obstacle and robotic formation size, respectively.

From the above discussion, it is clear that $J_{\text{trans}}$ hinders the deformation of the robot team, $J_{\text{pass}}$ limits the size of the formation, while $J_{\text{cross}}$ is the key item to ensure the obstacle crossing. With these cost functions, the optimization problem is formulated as 
\begin{subequations}
	\label{Jf}
	\begin{align}
 \min_{\mathcal{R}_{\Delta}} \;\, & J_f(\mathcal{R}_{\Delta})  \label{opt_Jf}\\
		 {\text{\hspace{-5mm} subject to}} \;\, &	\| \bm{p}_o - \bm{p}_i \| = \| \bm{v}_o -\bm{v}_i\|, \,  i \in \bm{I}_N,  \label{cons_Jtrans}\\
		 		& W \leq W_\text{convex}, z_{\text{obs}} \leq z_{\text{obsmax}}, {d}_{\text{obs}} \leq d_{\text{obsmax}},  \label{cons_Jpass} 
	\end{align}
\end{subequations}
where the indicators $z_{\text{obsmax}}$ and $d_{\text{obsmax}}$ of the robotic formation should be larger than the $z_{\text{obs}}$ and $ d_{\text{obs}}$ of the local obstacle. Constraint \eqref{cons_Jpass} enforces that the robot formation configuration variables defined in Section~\ref{sysconfig} are within their corresponding limits. 

\subsection{Local Motion Planning} 
\label{sec4_B}

With the optimized robot formation $\mathcal{R}_\Delta$, we still need to specify their trajectory $\mathcal{R}_N$ to pass through one obstacle at a time. We keep the robot formation and try to rotate the entire formation shape to find a trajectory $\bm{\tau}$ for crossing or bypassing obstacles.

\begin{figure}[h!]
	\hspace{-3.5mm}
	\centering
	\subfigure[]{
		\label{fig5a}
		\includegraphics[height=1.4 in]{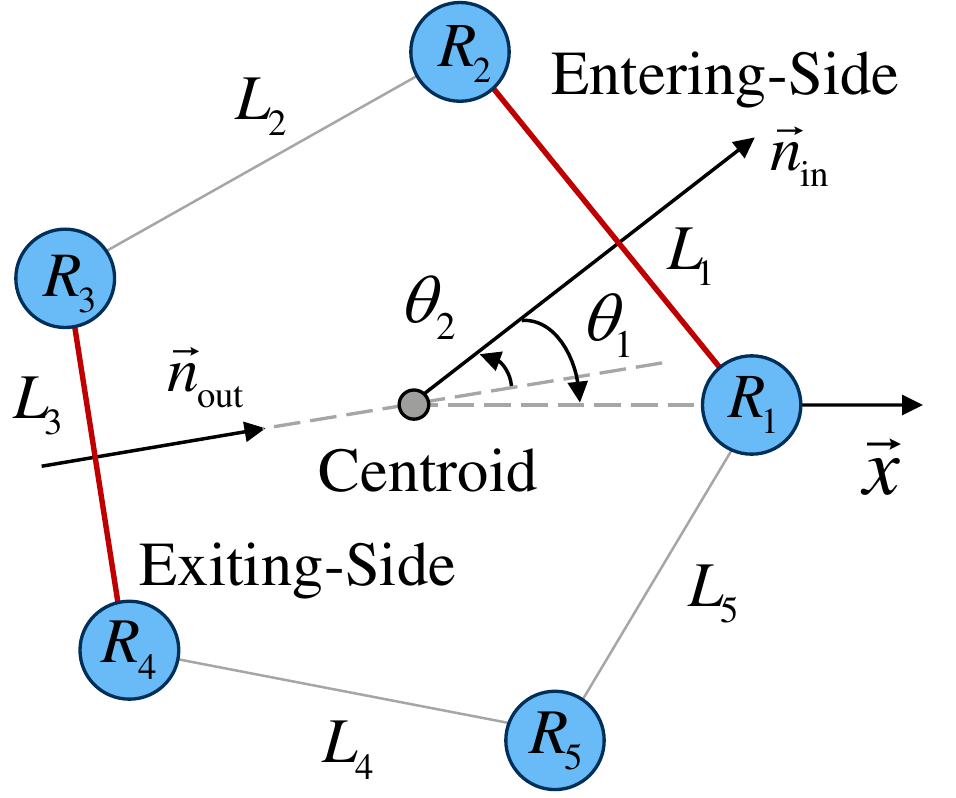}}
	\hspace{-3mm}
	\subfigure[]{
		\label{fig5b}
		\includegraphics[height=1.5 in]{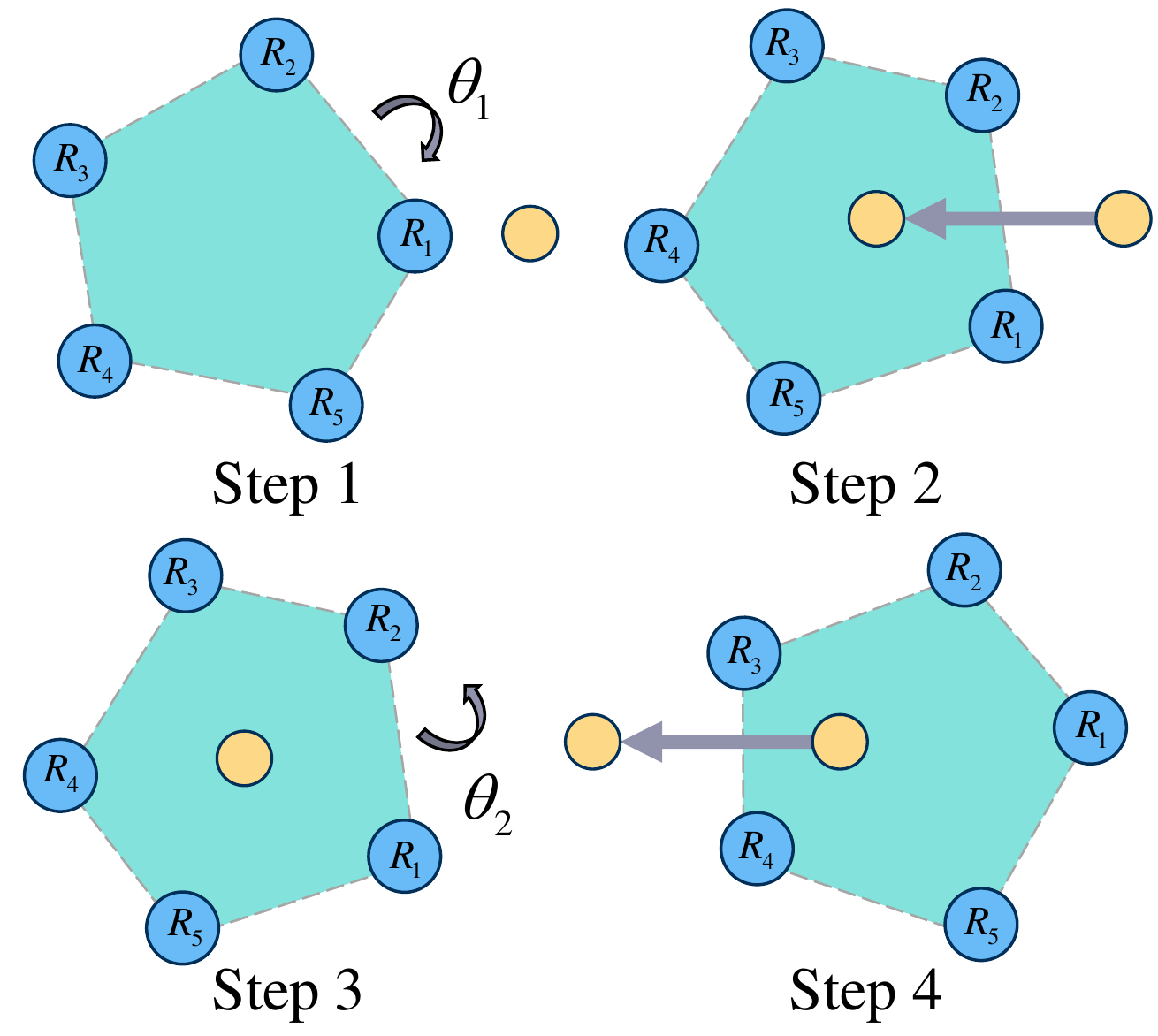}}
	\caption{Local motion planning for formation crossing the obstacle. (a) The entering and exiting sides of the formation when crossing the obstacle. (b) The four steps of movement after the robot formation enters the obstacle influence region: rotation, translation, rotation, translation.
	}  \label{fig_path}     
\end{figure}

 Fig.~\ref{fig_path} illustrates the robot path generation design for a five-robot team. As shown in Fig.~\ref{fig5a}, each side of the robot formation $\mathcal{R}_\Delta$ is denoted as $L_i$. We consider the forward velocity direction of the formation centroid as the $x$-axis. By the definition of $d_{\text{obsmax}}$ in~\eqref{eq_zobs}, the robot formation can pass from the top of the obstacle from any $L_i$ and the normal vectors for the obstacle to enter or exit the formation are denoted as $\overrightarrow{\bm{n}}_\text{in}$ and $\overrightarrow{\bm{n}}_\text{out}$, respectively. There are four steps for obstacle crossing, as shown in Fig.~\ref{fig5b}. In the first step, $t \in (0, T_1]$, the robot formation rotates about its centroid for angle $\theta_1$ such that  $\overrightarrow{\bm{n}}_\text{in}$ coincides with the $x$-axis. The formation then translates to cross the obstacle from time moments $T_1$ to $T_2$. In the third step, $t \in (T_2, T_3]$, the formation rotates with angle $\theta_2$ and vector $\overrightarrow{\bm{n}}_\text{out}$ aligns with the $x$-axis, where $\theta_2$ is the relative angle between $\overrightarrow{\bm{n}}_\text{in}$ and $\overrightarrow{\bm{n}}_\text{out}$; see Fig.~\ref{fig5a}. Finally, the formation travels to pass from the top of the obstacle at time $T_4$. During the entire process, the translational speed $v$ of the robot formation is assumed to be constant, and the time instances $T_i$, $i=1,\ldots,4$, are determined by the moving distance.

We denote the robot formation centroid path as $\bm{\pi}(t)=(x_{\tau}(t),y_{\tau}(t),\theta_{\tau}(t))$. From the above discussion, we have  $y_{{\tau}}(t)=0$ and a path generation is then designed as 
\begin{equation*}
(x_{\tau}(t),\theta_{\tau}(t))=\begin{cases} \left(0,\frac{\theta_1}{T_1}t\right) &  t\in (0,T_1],\\
	 (vt,\theta_1) & t\in (T_1,T_2], \\
	 \left(v\delta_T,\frac{\theta_2(t-T_2)}{T_3 - T_2} +\theta_1\right) &  t\in (T_2,T_3], \\
	 (v(t+\delta_T-T_3), \theta_1 + \theta_2) & t\in (T_3,T_4],
	 \end{cases}
	\label{eq_tau}
\end{equation*}
where $\delta_T=T_2-T_1$. $\theta_1$ and $\theta_2$ are determined according to different entering and exiting side, which are positive in the counterclockwise direction. Algorithm~\ref{Al_LPG} summarizes the path generation design for the robot team. Compared with the existing obstacle bypassing paths, the path generated by Algorithm~\ref{Al_LPG} is to let the transported object cross over the obstacle.

\begin{algorithm}[ht!]
	\vspace{1mm}
	\SetAlgoVlined
	\label{Al_LPG}
	\caption {Local Path Generation}
	\SetKwInOut{Input}{Input}
	\SetKwInOut{Output}{Output}
	\SetKwFunction{Linecross}{Linecross}
	\SetKwFunction{Polygon}{Polygon}
	\Input{$\mathcal{R}_\Delta$, $z_{\text{obs}}$, $d_{\text{obs}}$, $W_{\text{convex}}$}
	\Output{$\bm{\pi}(t)$ and $\bm{r}_i(t),\, i=1,\ldots,N$.}
	
	\eIf{$\textit{CrossingFormationExist} (\mathcal{R}_\Delta, z_{\text{obs}}, d_{\text{obs}}, W_{\text{convex}})$ }{	
		$\bm{\pi}(t)$  $\gets$ $\textit{CrossingPathGeneration}$ ($\bm{\tau}$)	\\
		$\bm{r}_i(t) \gets \textit{FormationToRobots}(\bm{\pi}(t))$ \\  		
		\Return $\bm{r}_i(t)$
	}{
		  \eIf{$\textit{BypassingFormationExist} (\mathcal{R}_\Delta, z_{\text{obs}}, d_{\text{obs}}, W_{\text{convex}})$}{
		  	$\bm{\pi}(t)$  $\gets$ $\textit{BypassingPathGeneration}$ ($\bm{\tau}$) \\
		  	$\bm{r}_i(t) \gets \textit{FormationToRobots}(\bm{\pi}(t))$  \\
		  	\Return $\bm{r}_i(t)$
	  }{
	  		\Return  $\textit{FALSE}$
      }
	} 
\end{algorithm}

\section{Experimental Results}
\label{sec5}

In this section, we demonstrate the planning algorithms by experiments and simulation. A three-robot team is used as an experimental platform and simulation demonstrates the motion planning algorithms for more than three robots case.

\vspace{0mm}
\subsection{Experimental Setup}

\begin{figure}[h!] 
	\includegraphics[width=0.95\columnwidth]{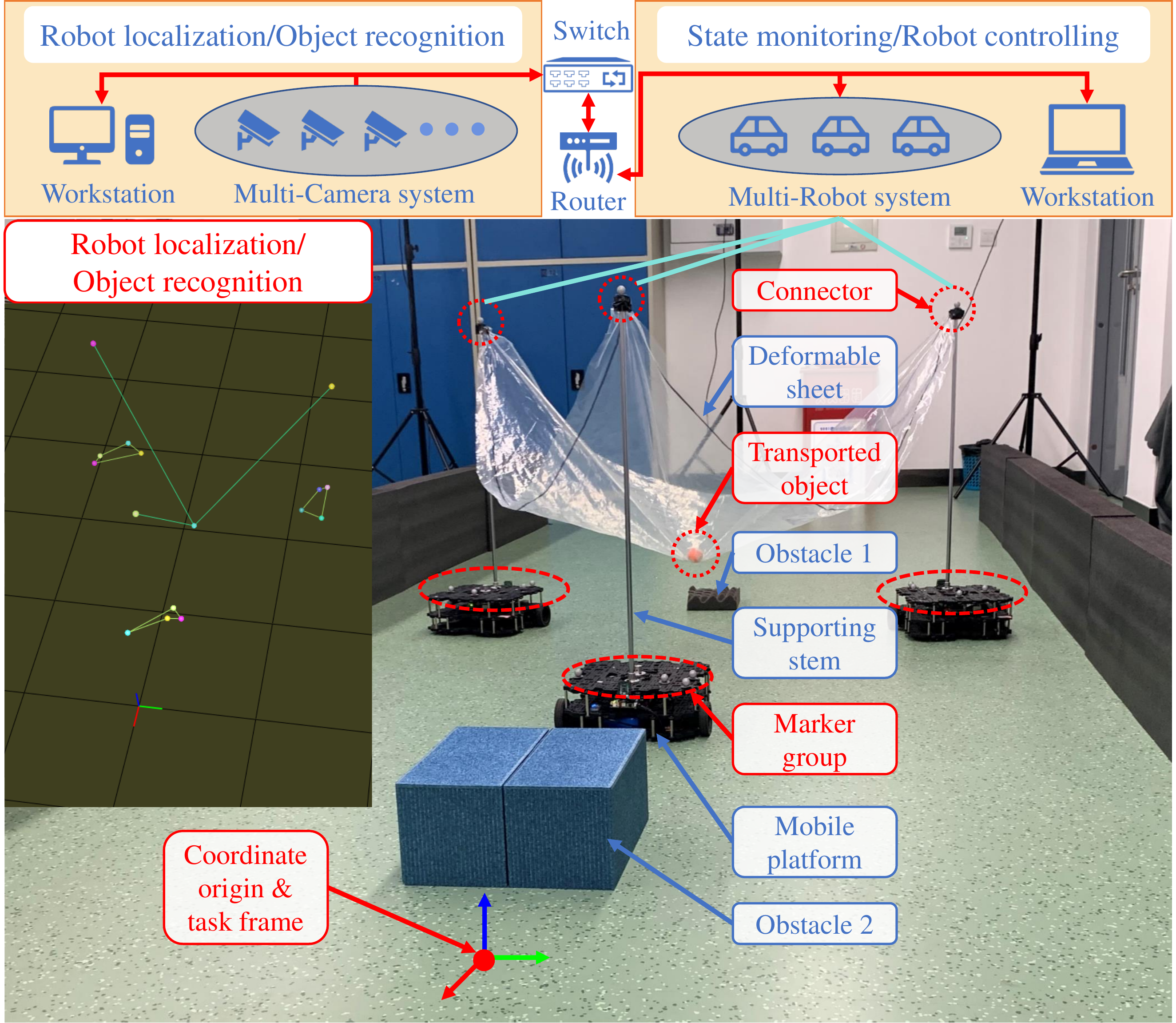}
	\centering
	\caption{Experimental setup and communication network of the multi-mobile robot transportation system.}
	\label{fig_exp_setup}       
	\vspace{-0mm}
\end{figure}

Fig.~\ref{fig_exp_setup} shows the experimental setup and communication network of the system. Three robots (Turtlebot3 Waffle-Pi) were used in experiment. The transported object was a billiard ball. The position of the robots and the object were acquired by a motion capture system (8 cameras from NOKOV) at a rate of $60$~Hz. Plastic cloth sheets were used in experiments. Each side length of the sheet was $1.6$~m and the height of each holding point was $z_r=0.79$~m. The experimental environment was a corridor (2 meters wide, 6 meters long) with two different-height obstacles in the middle. The first obstacle (Obstacle \#1) was a circular shape with a radius of $0.1$~m and a height of $5$~cm, and the other (Obstacle \#2) had a radius of $0.2$~m and a height of $0.2$~m. The robot controller was implemented in robot operating system (ROS) with embedded system (NVIDIA Jetson NANO and OpenCR controller). A remote laptop (Intel i7-4710HQ CPU with four cores and 8 GB RAM) was set up for state monitoring and robot controlling. The following model parameters were used in experiments: $\lambda_1=1$, $\lambda_2=1$, $\lambda_3=1$, $\lambda_4=10$, $\lambda_5=10$, $z_{\text{safe}}=4$ cm, and $\Delta r=5$ cm.

\begin{figure*}[th!]
	\vspace{-3mm}
	\hspace{-3.5mm}
	\centering
	\subfigure[]{
		\label{fig7a}
		\includegraphics[width=1.8in]{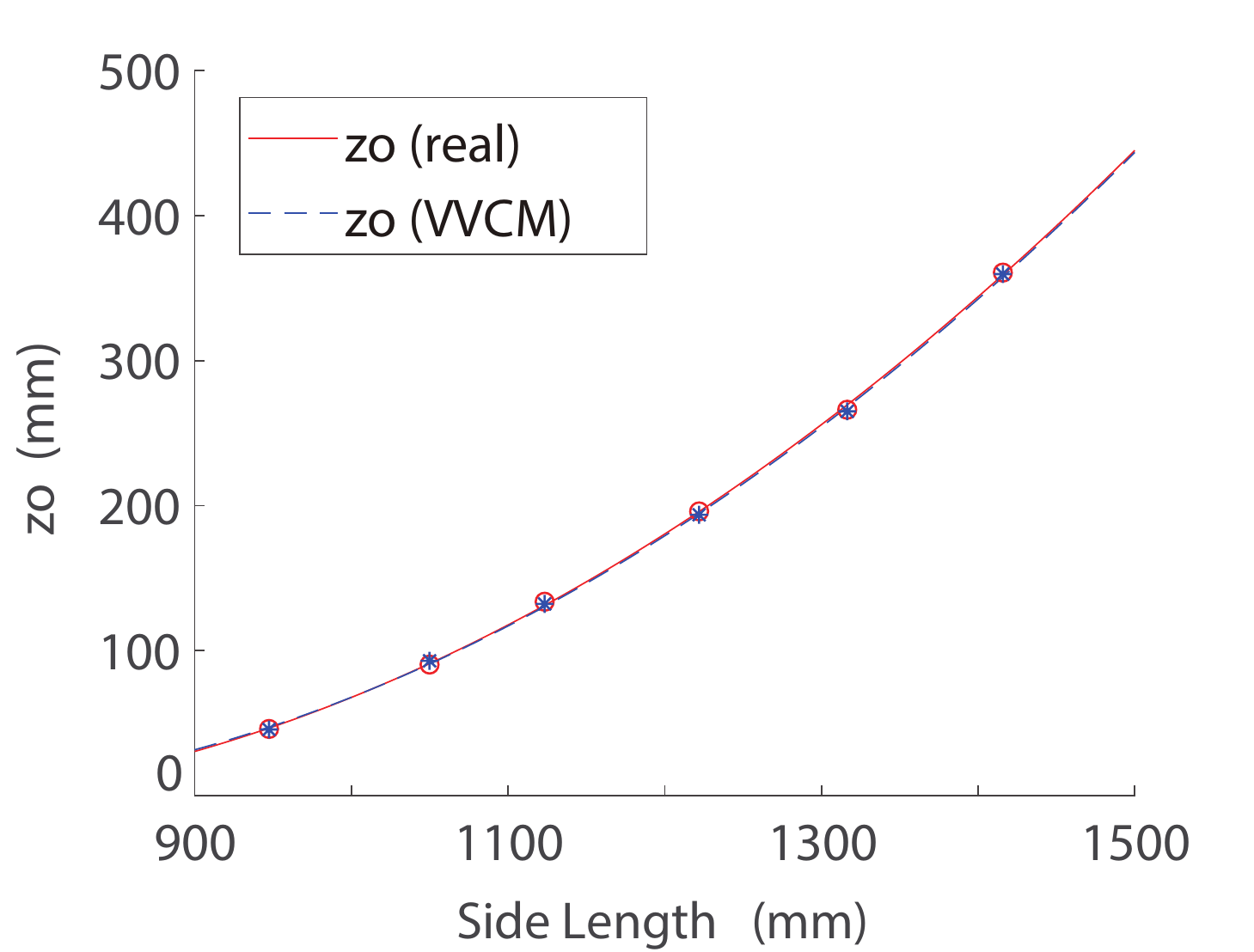}}
	\hspace{-3mm}
	\subfigure[]{
		\label{fig7b}
		\includegraphics[width=1.8in]{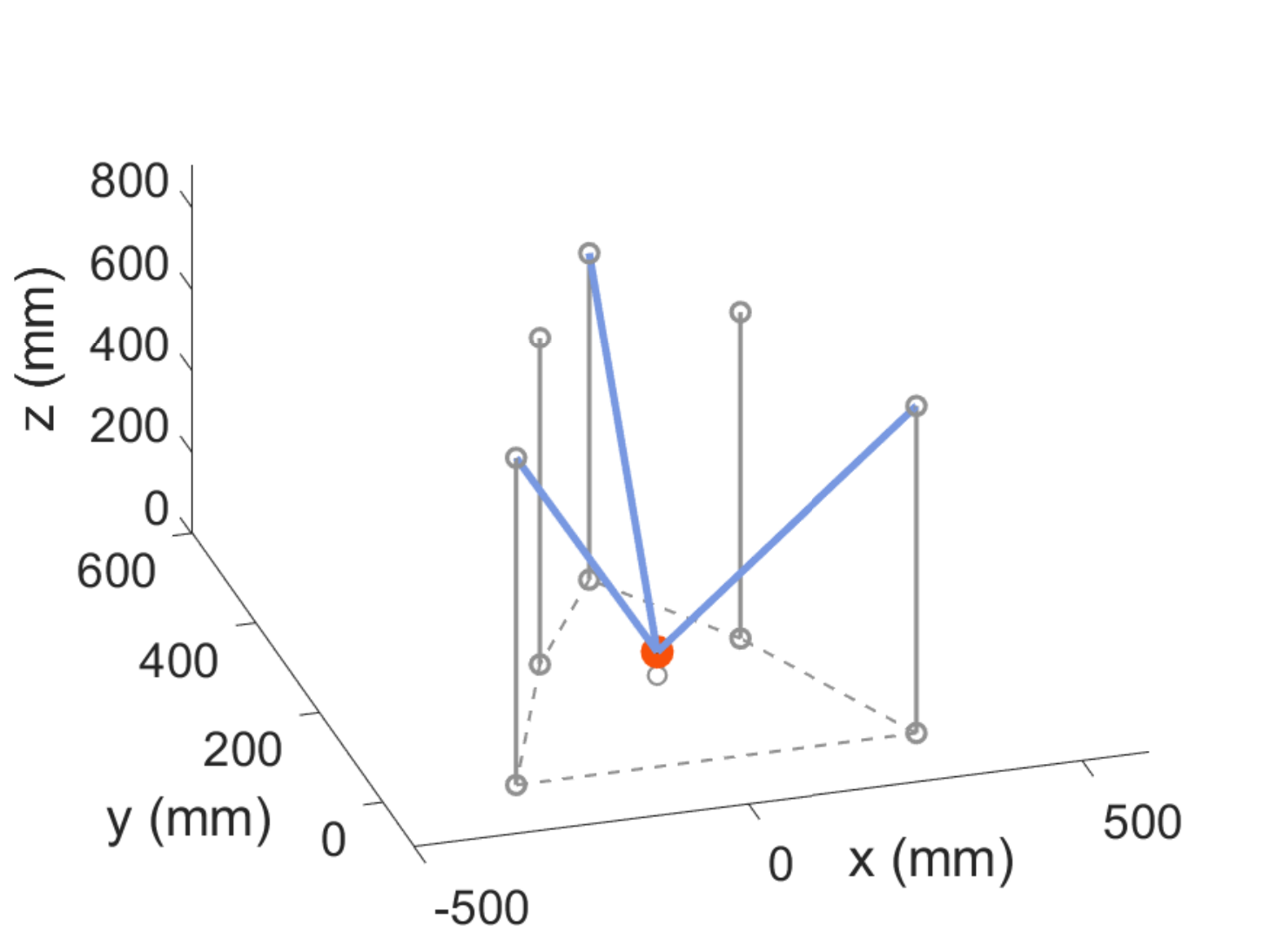}}
	\hspace{-4mm}
	\subfigure[]{
		\label{fig7c}
		\includegraphics[width=1.8in]{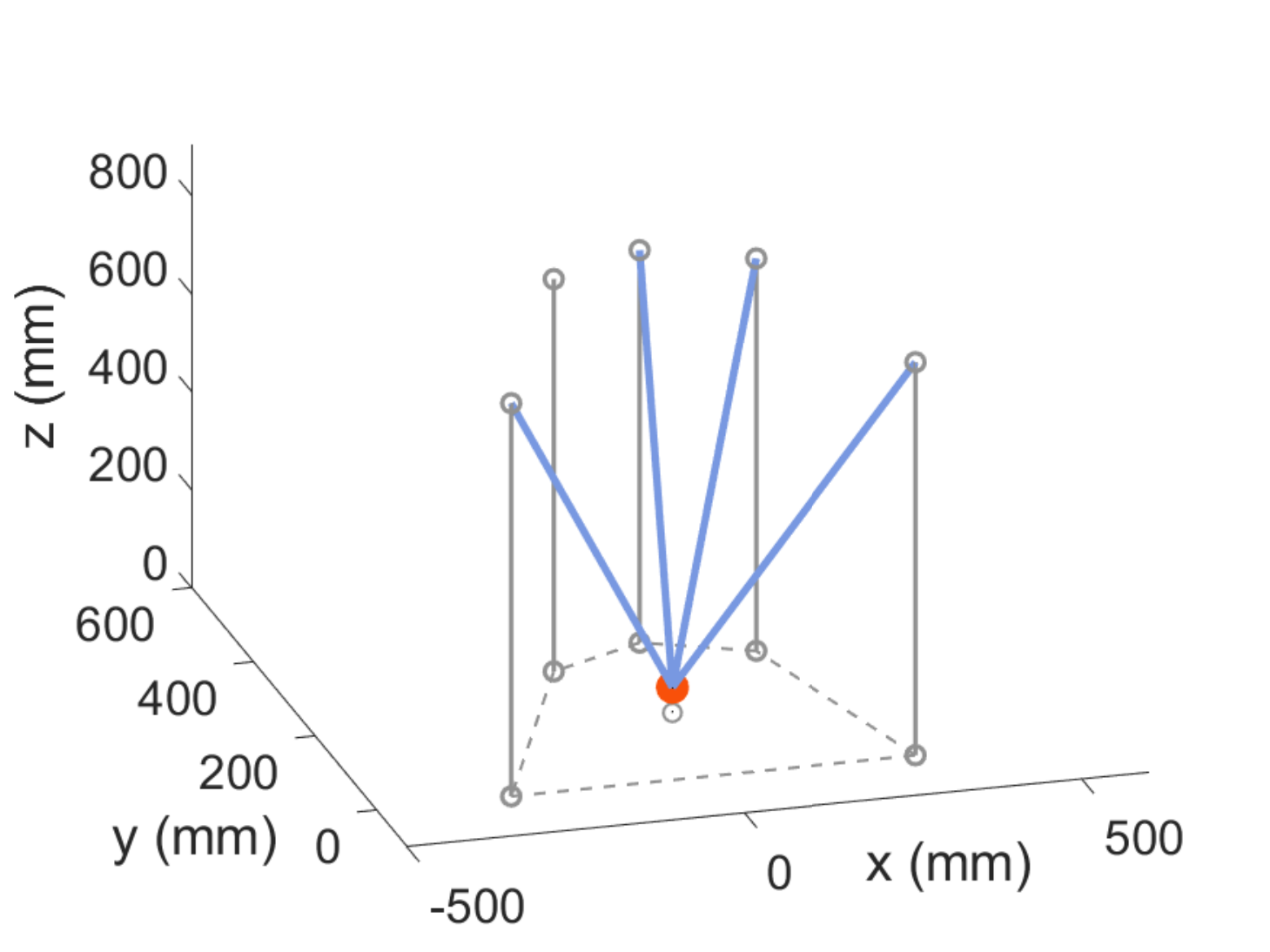}}
	\hspace{-4mm}
	\subfigure[]{
		\label{fig7d}
		\includegraphics[width=1.8in]{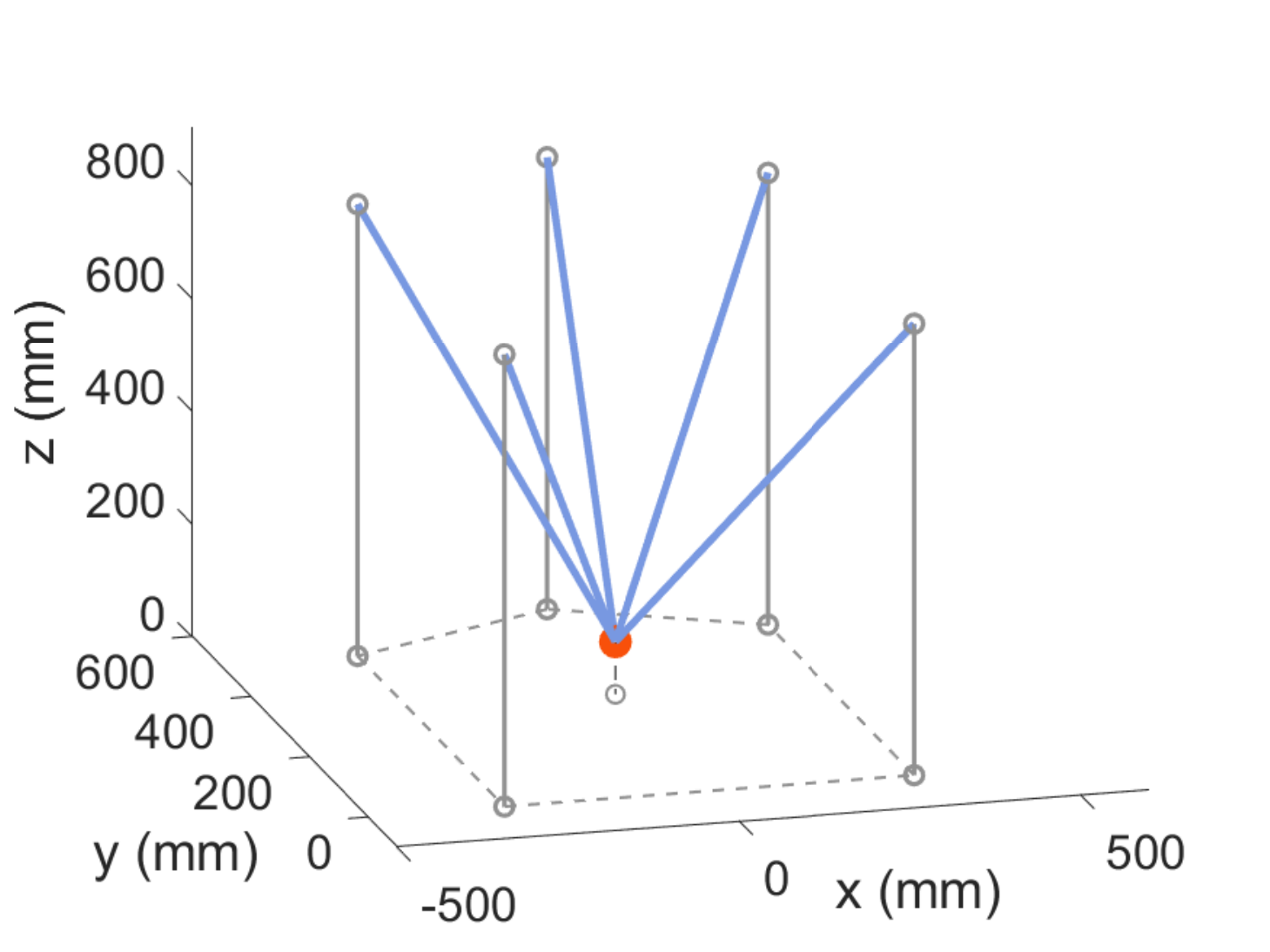}}
	\hspace{-4mm}
	\caption{(a) The actual and VVCM-calculated heights of the object in an equilateral triangle formation and their fitted curves. (b)-(d) show the VVCM calculation examples when three, four, and five virtual cables are taut, respectively.}
	\label{fig_exp_vvcm}       
\end{figure*}

\begin{figure*}[th!]
	\includegraphics[width=\linewidth]{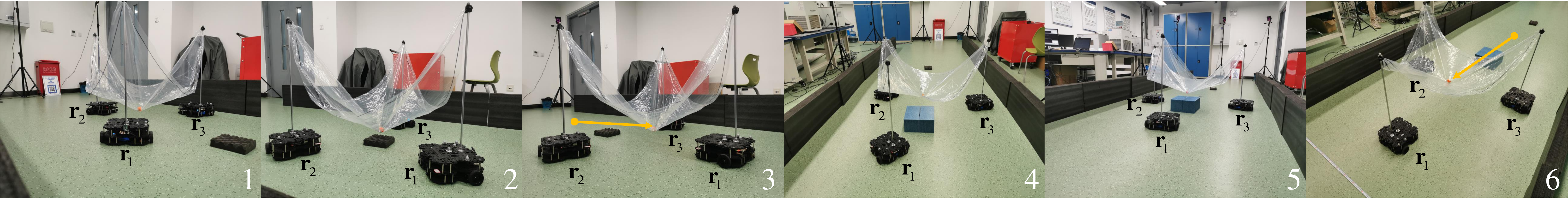}
	\centering
	\caption{The snapshots of the three-robot team that carried and transported a billiard ball with a deformable sheet. Two obstacles different heights are placed on the trajectory of the robotic team.}
	\label{fig_exp_3}       
\end{figure*}

\begin{figure*}[th!]
	\vspace{-3mm}
	\hspace{-8mm}
	\subfigure[]{
		\label{fig9a}
		\includegraphics[width=2.63in]{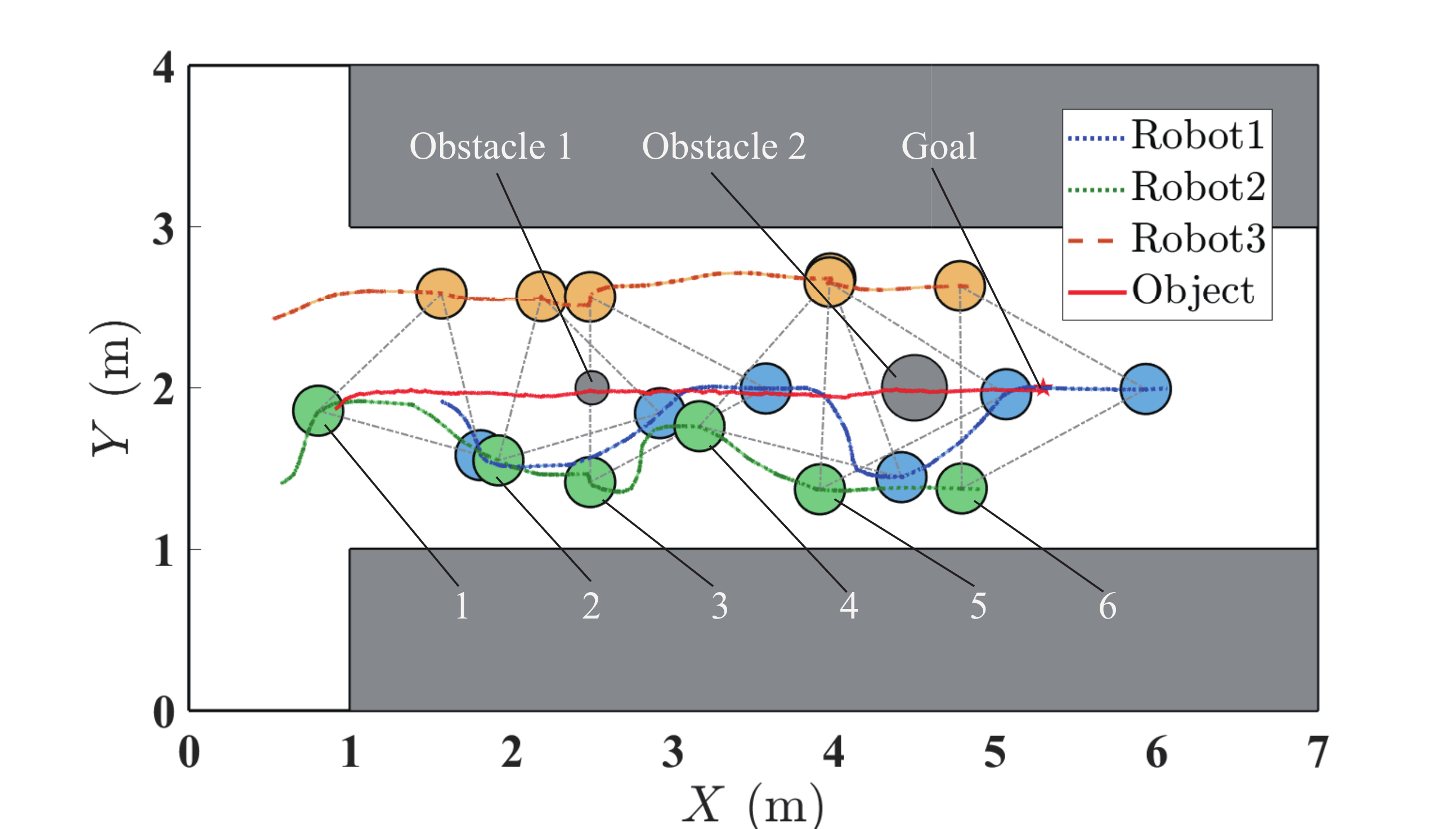}}
	\hspace{-9mm}
	\subfigure[]{
		\label{fig9b}
		\includegraphics[width=1.7in]{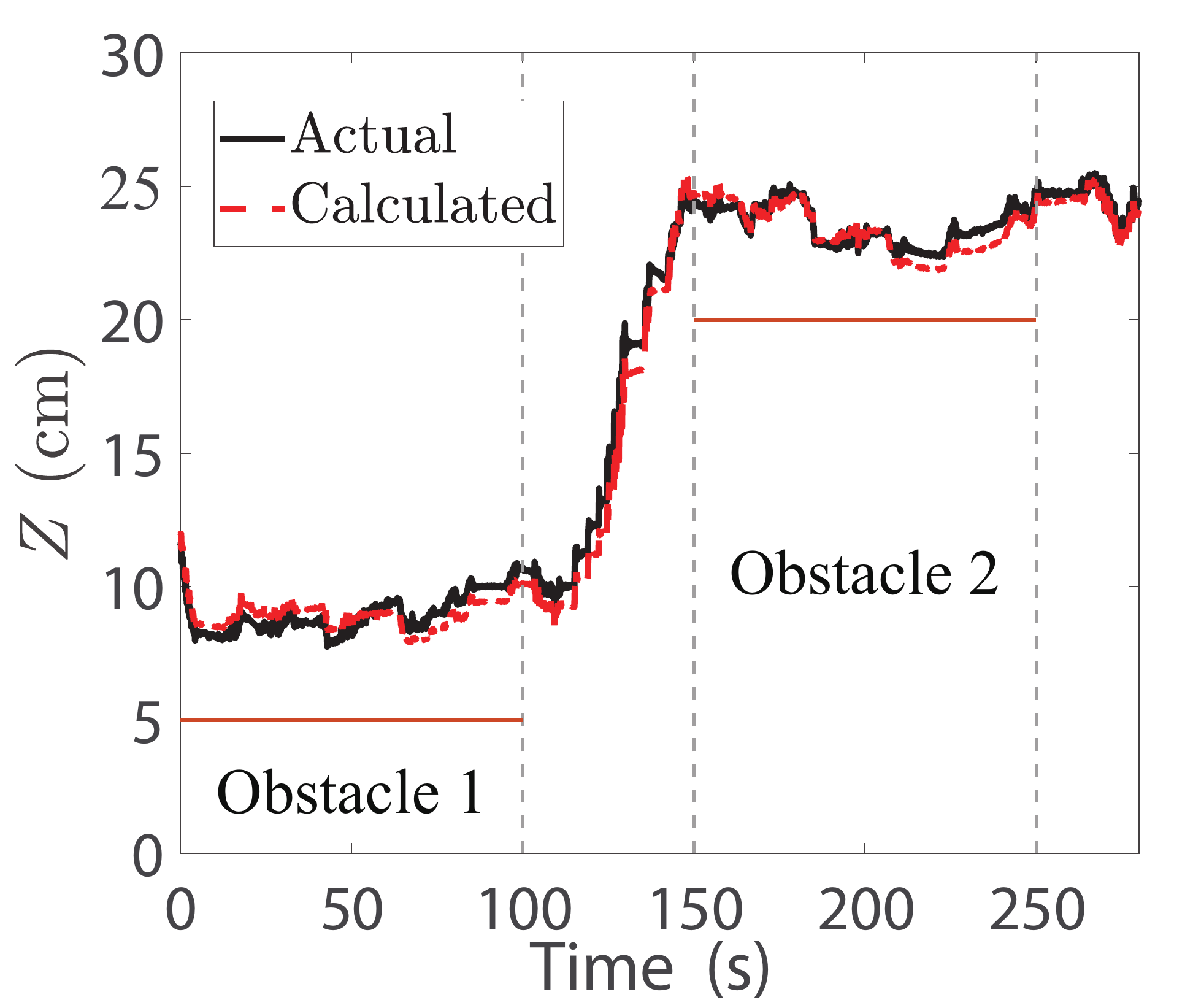}}
	\hspace{-5mm}
	\subfigure[]{
		\label{fig9c}
		\includegraphics[width=1.55in]{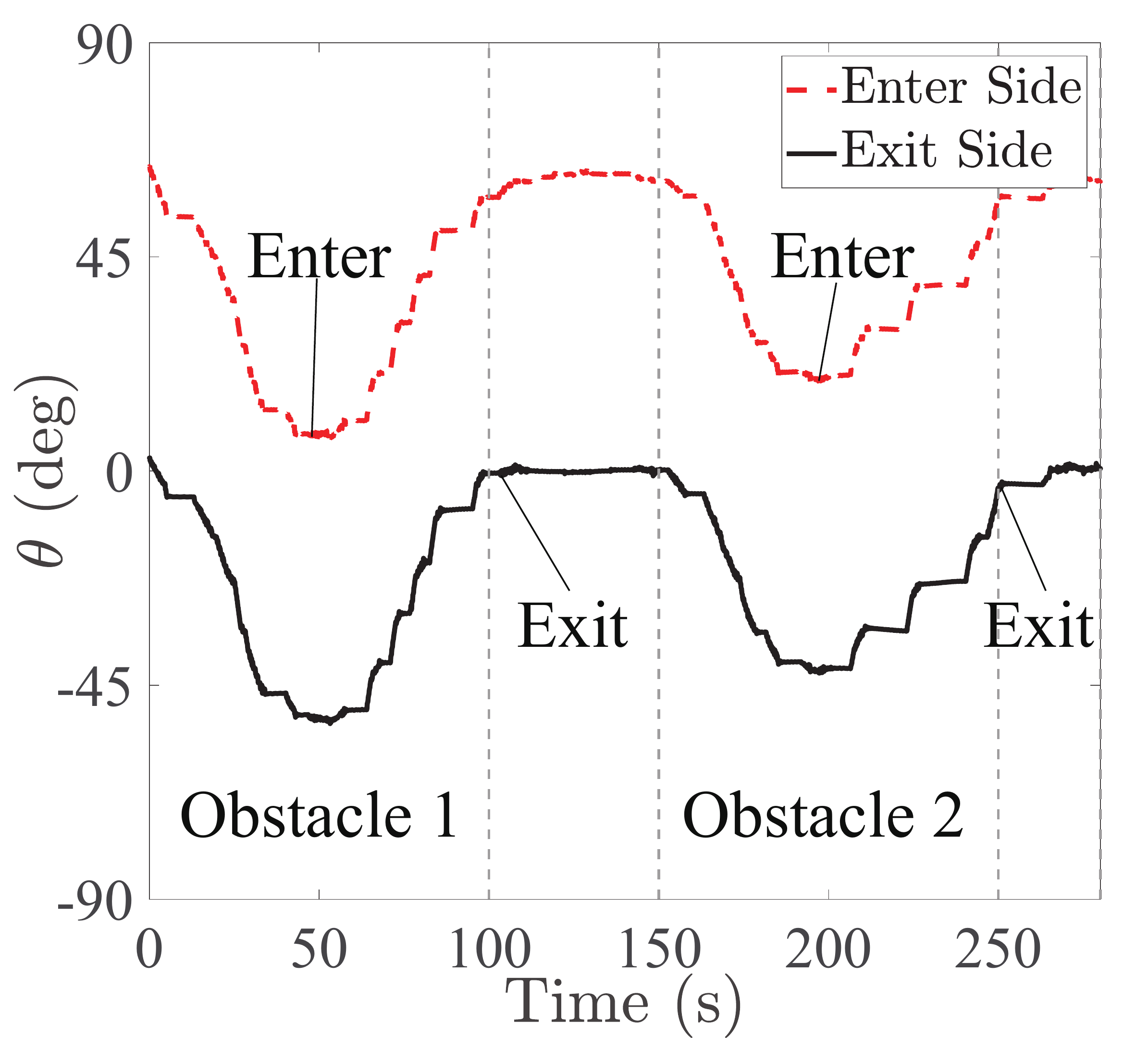}}
	\hspace{-4mm}
	\subfigure[]{
		\label{fig9d}
		\includegraphics[width=1.67in]{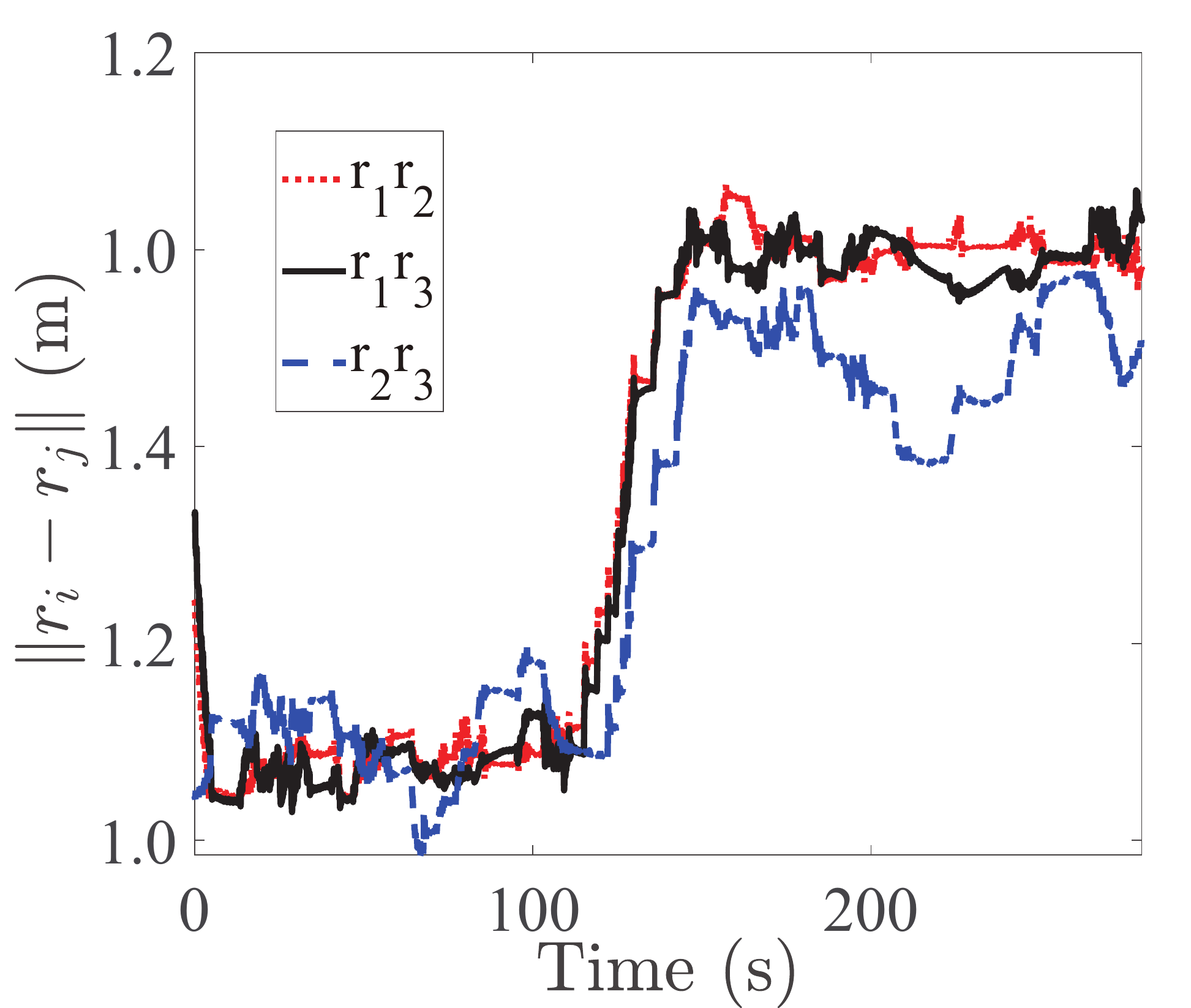}}
	\caption{Experimental results with the object transported by a three-robot team. (a) Trajectory of three robots (dashed lines) and the object (red line). The robot goal (ending point) is represented by a red star marker. The labeled numbers correspond to the snapshots shown in Fig.~\ref{fig_exp_3}. (b) The height trajectory $z_o$ of the object during the transporting. Two-obstacle heights are marked as straight-line. (c) The orientation angle $\theta$ of the normal vectors of the entering and exiting sides of robot formation. (d) The relative distances $\|\Delta \bm{r}_{ij}\|$ among the robots.}
	\label{fig_exp_3_real}       
	\vspace{-0mm}
\end{figure*}

\vspace{-3mm}
\subsection{Experimental Results}

We first show the effectiveness of the VVCM and Algorithm~\ref{Al_DK}. The accuracy of the object height is the key to verifying the validity of the model. The first experiment was conducted with the equilateral triangle formation as an example to verify the model. The actual heights of the object under different equilateral triangle formations were measured and compared with the VVCM. Fig.~\ref{fig_exp_vvcm}(a) shows the validation of the object height prediction by the VVCM. The root mean squared error (RMSE) of the height prediction by the VVCM is $15.7$~mm, and the error in the $XY$ plane is $16.2$~mm. The model prediction error might mainly come from the positioning measurement variation of the billiard ball. The results confirm that the VVCM prediction is consistent with the experiments under different formation variations. Figs.~\ref{fig_exp_vvcm}(b)-(d) show the VVCM calculation examples of direct kinematics under different combinations of taut/slack status of virtual cables in a five-robot team. These results confirm that the position of the object can be obtained by the VVCM for any feasible formation. 

Due to the narrow corridor environment with obstacles in experiments, the motion planner such as that in~\cite{alonso2017multi} cannot obtain the feasible trajectory successfully because it only considers the obstacle bypassing in 2D space. The proposed motion planner instead generated the feasible obstacle-crossing trajectory successfully. Fig.~\ref{fig_exp_3} shows the snapshots of the robot formation and trajectory and Fig.~\ref{fig9a} shows the detailed trajectories of the robot motion. The object went straight crossing over the two obstacles and the RMSE in the $XY$ plane is $31.1$~mm. For crossing Obstacle \#1, the average side lengths of the robot formation were $1.04$, $1.04$ and $1.05$~m (exiting side). For crossing Obstacle \#2, the average side lengths of the formation were $1.30$, $1.29$ and $1.24$~m (exiting side). Fig.~\ref{fig9b} shows the comparison results of the height of the object  in experiments and the VVCM prediction. The RMSE of the VVCM prediction is $5.3$~mm. The average heights of the object in the process of crossing Obstacles \#1 and \#2 were respectively $9.0$~cm and $23.4$~cm, which were both safe. Fig.~\ref{fig9c} shows the angles of the normal vectors at the entering and exiting sides. The entering angles for the robot formation to cross two obstacles were $6$ and $18^{\circ}$, respectively, and the exiting angles were both $0^{\circ}$, which was consistent with the local path generation method discussed in Section~\ref{sec3_B}. Fig.~\ref{fig9d} shows the relative distances among the robots. It is clear that the relative distance changes corresponded to the object's height changes as shown in Fig.~\ref{fig9b}. The main source of the fluctuation in these figures was from the robot speed variations due to the nonholonomic nature of the used robots when the formation rotated.

\begin{figure*}[th!]
	\vspace{-5mm}
	\hspace{-2mm}
	\subfigure[]{
		\label{fig10a}
		\includegraphics[height=1.65in]{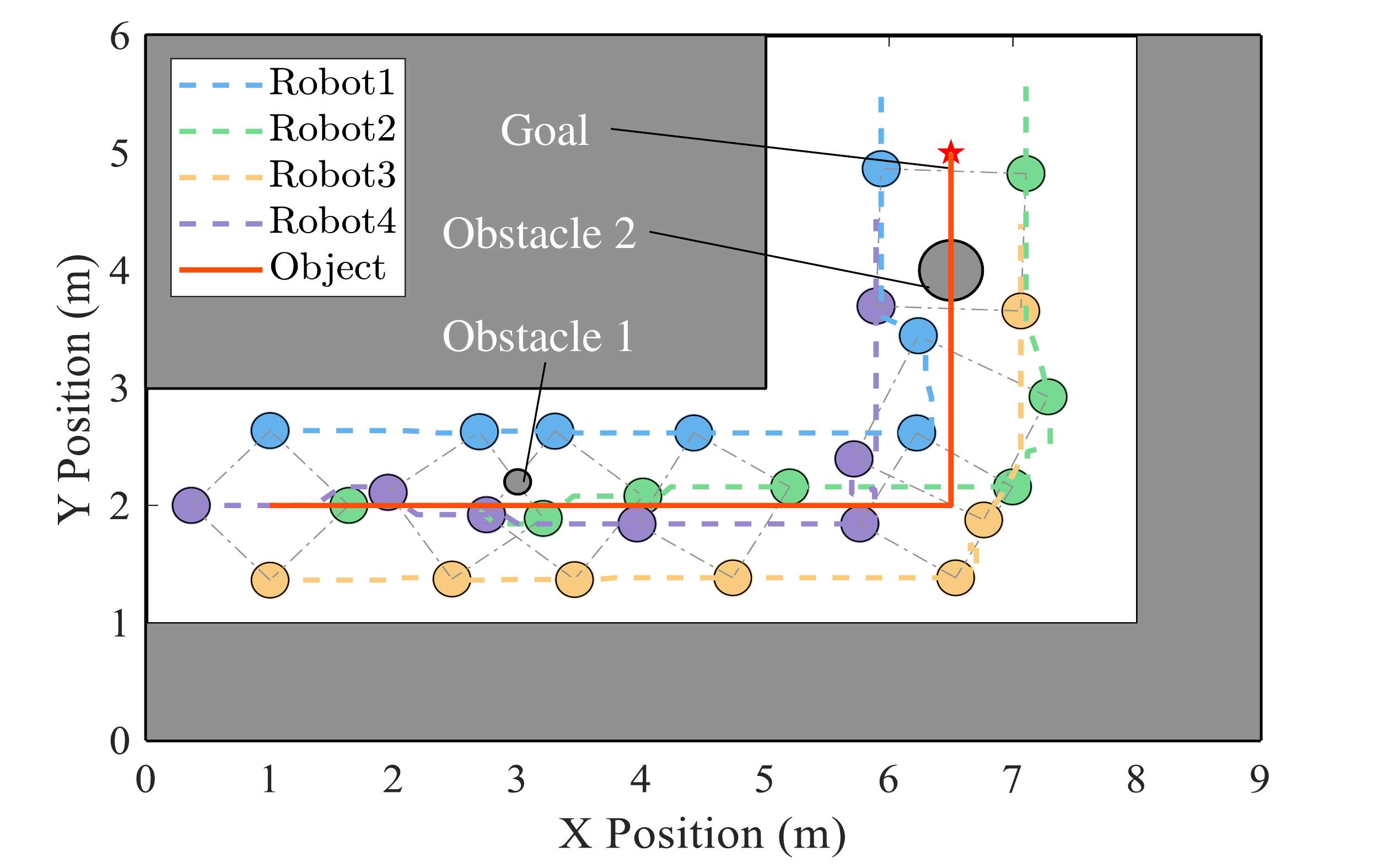}}
	\hspace{-6mm}
	\subfigure[]{
		\label{fig10b}
		\includegraphics[height=1.65in]{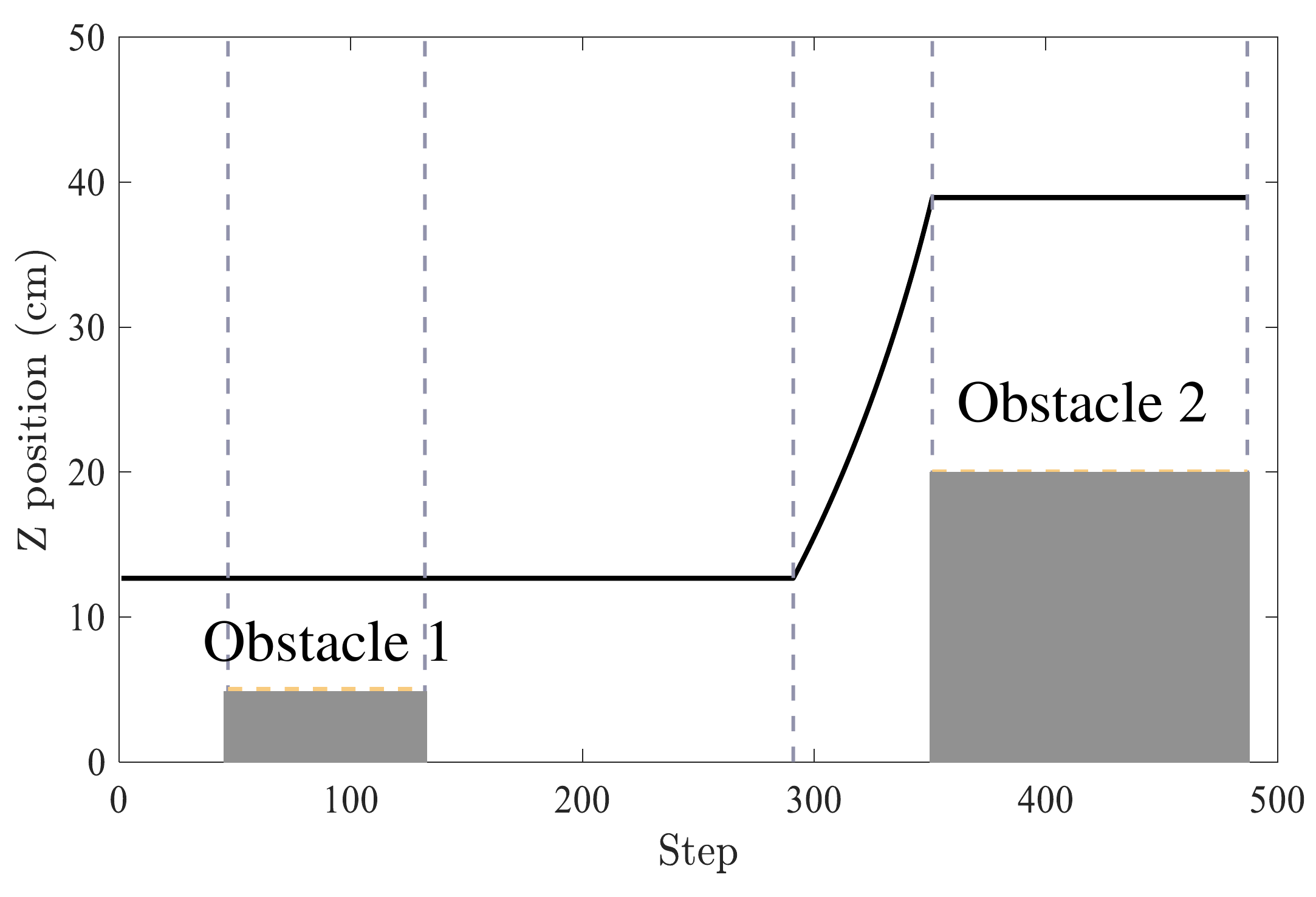}}
	\hspace{-4mm}
	\subfigure[]{
		\label{fig10c}
		\includegraphics[height=1.75in]{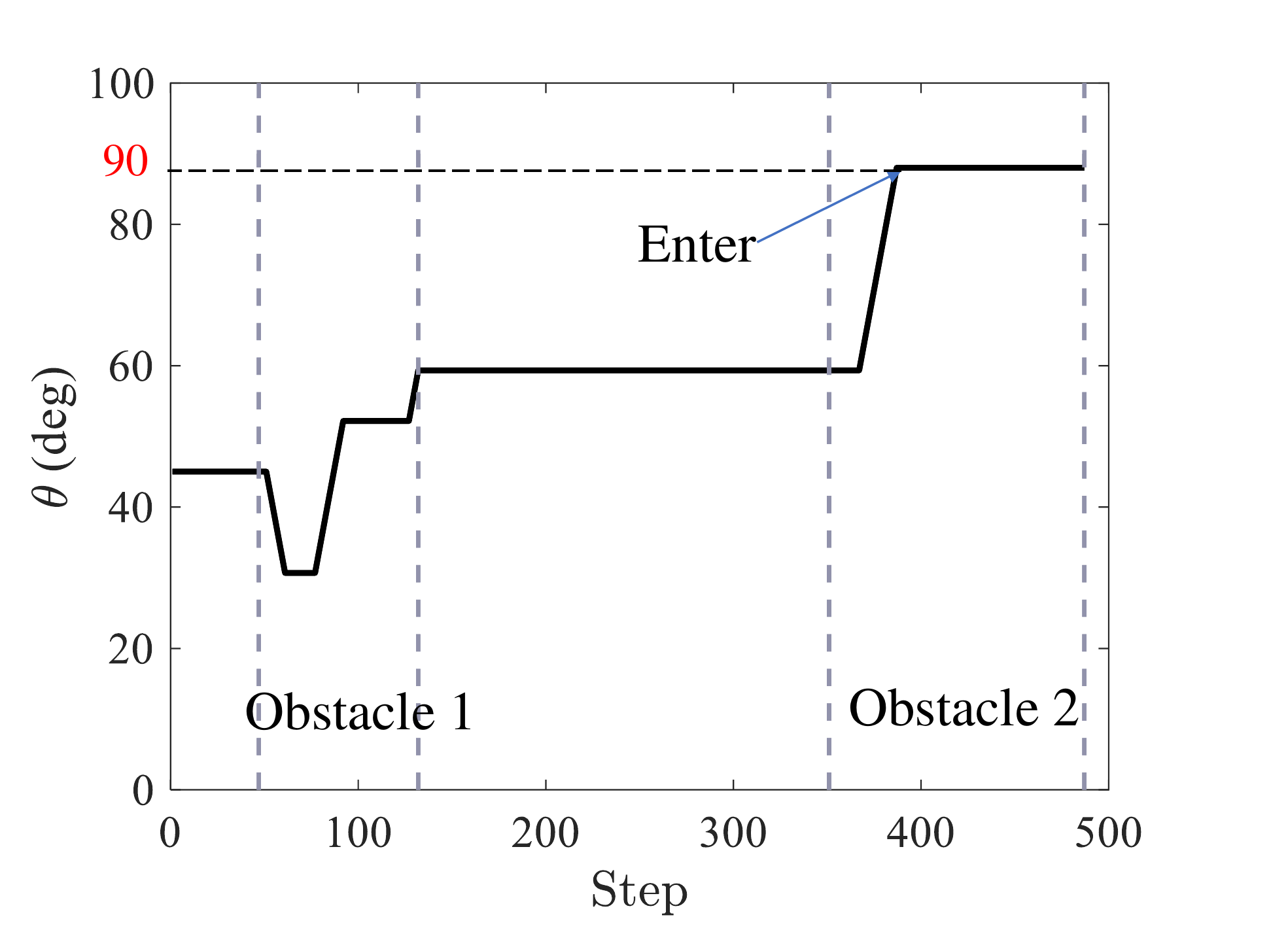}}
	\caption{Simulation results of a $4$-robot team navigating in a turned corridor. (a) Trajectory profiles of the $4$-robot team (dashed lines) and the transported object (red line). The ending point for the object is represented by the star marker. (b) The height of the object during transporting and its relationship with the two obstacles. (c) The normal vector orientation profiles of the entering side, which finally coincides with the center-line of the vertical corridor.}
	\label{fig_exp_4_sim}       
	\vspace{-0mm}
\end{figure*}

We further run simulation studies with $4$- and $6$-robot teams in different scenarios to verify the scalability of the motion planner. Fig.~\ref{fig_exp_4_sim} shows the results for a $4$-robot team to pass a turned corridor while carrying the object. Fig.~\ref{fig10a} shows the robot motion trajectories, Fig.~\ref{fig10b} shows the object height trajectory during transporting and Fig.~\ref{fig10c} demonstrates the entering angle profiles of the robot formation when crossing obstacle. In this example, different exiting edges of the robot formation were selected and when there is no need to rotate, the formation can pass the obstacle from the top directly. The robot formation successfully followed the center-line of the trajectory, while rotating was used to pass crossing two obstacles. Indeed, the planner can be widely used to various numbers of robots and the algorithm is scalable. To apply this planner to complex environments, the robots should be equipped with diverse sensors to detect obstacles and build the map in real time. Distributed sensing and real-time control algorithms are required in these applications. Due to page limit, we demonstrate the detailed results of a $6$-robot team in the companion video clip.

\section{Conclusion} 
\label{sec6}

We proposed to use the virtual variable cables model to capture motion of the object in the deformable sheet that was  collaboratively held by a multi-robot transporting system. Based on the VVCM, a motion planner was then proposed for transporting the object by the multi-robot. One attractive feature of the motion planner was its capability for crossing obstacles in 3D space. Compared with existing collaborative multi-robot object transporting approaches, the proposed planner gave priority to crossing obstacles instead of bypassing obstacles. This property improved the efficiency to traverse through cluttered environments (e.g., narrow corridors with obstacles) in which the traditional methods cannot achieve. The experiments and simulation were also presented to validate the VVCM and demonstrate the planning algorithms. 

\bibliographystyle{IEEEtran}
\bibliography{References}

\end{document}